\documentclass[11pt]{article}

\usepackage[margin=1in]{geometry}
\usepackage{iftex}
\usepackage{microtype}
\usepackage{url}
\usepackage[numbers,sort&compress]{natbib}
\usepackage{amsmath,amssymb}
\usepackage{mathtools}
\usepackage{booktabs}
\usepackage{tabularx}
\usepackage{array}
\usepackage{longtable}
\usepackage{graphicx}
\usepackage{pifont}
\usepackage{xcolor}
\usepackage[most]{tcolorbox}
\usepackage{fvextra}
\usepackage{ragged2e}
\usepackage{subcaption}
\usepackage{pdflscape}
\usepackage{multirow}
\usepackage{seqsplit}
\usepackage{hyperref}

\definecolor{mygray}{HTML}{4D4D50}
\definecolor{myblue}{HTML}{7EA6E0}
\definecolor{darkblue}{rgb}{0,0,0.5}
\definecolor{editred}{RGB}{0,0,0}
\definecolor{tablegreen}{HTML}{1B8A3B}
\definecolor{tablered}{HTML}{C62828}

\newcommand{\cmark}{\textcolor{tablegreen}{\ding{51}}}
\newcommand{\xmark}{\textcolor{tablered}{\ding{55}}}

\hypersetup{
    colorlinks=true,
    citecolor=darkblue,
    linkcolor=darkblue,
    urlcolor=darkblue,
    breaklinks=true
}

\title{\textbf{AuthBench: A Large-Scale Multilingual Benchmark for\\
Authorship Representation across Genres and Lengths}}

\author{
MaoXun Huang \qquad
Zhenxing Zhang \qquad
Claire Cardie
\\[0.5em]
Department of Computer Science, Cornell University\\
Ithaca, NY, United States
}

\newcommand{\rev}[1]{#1}
\newcommand{\rededit}[1]{#1}
\newcommand{\ind}[1]{\mathbf{1}\!\left[#1\right]}

\newcolumntype{L}[1]{>{\raggedright\arraybackslash}p{#1}}
\newcolumntype{Y}{>{\RaggedRight\arraybackslash}X}
\graphicspath{{./}}
\newcommand{\safeincludegraphics}[2][]{%
  \IfFileExists{#2}{%
    \includegraphics[#1]{#2}%
  }{%
    \fbox{%
      \parbox[c][0.18\textheight][c]{0.9\linewidth}{%
        \centering Figure file not found\\\texttt{\detokenize{#2}}%
      }%
    }%
  }%
}

\ifPDFTeX
  \PackageError{colm_latex}{This manuscript requires XeLaTeX or LuaLaTeX because it contains CJK text}{Switch the Overleaf compiler to XeLaTeX.}
\else
  \usepackage{fontspec}
  \usepackage{xeCJK}
  \setCJKmonofont{FandolFang-Regular.otf}
\fi

\DefineVerbatimEnvironment{jsonblock}{Verbatim}{
  breaklines=true,
  breaksymbolleft=\tiny,
  fontsize=\small
}

\DefineVerbatimEnvironment{JSONVerbatim}{Verbatim}{
  breaklines=true,
  breakanywhere=true,
  breaksymbolleft=\raisebox{0.2ex}{\tiny$\hookrightarrow$}\,,
  fontsize=\footnotesize,
  baselinestretch=1,
  commandchars=\\\{\}, 
}

\newtcolorbox{jsonexample}[2][]{%
  enhanced,
  sharp corners,
  boxrule=0.4pt,
  colback=black!2,
  colframe=black!20,
  left=6pt,right=6pt,top=6pt,bottom=6pt,
  title={#2},
  fonttitle=\bfseries\small,
  attach title to upper,
  #1
}

\begin{document}
\maketitle

\begin{abstract}
Authorship signals matter in settings where writing style carries identity: digital forensics, plagiarism analysis, account linking, misinformation investigation, and machine-generated text detection. Yet current authorship benchmarks remain fragmented, usually covering only a narrow language set, a single genre, or a limited document-length regime, which makes it difficult to assess whether modern representations truly generalize. We introduce AuthBench, a large-scale multilingual benchmark for authorship representation that is designed to make this evaluation broad, standardized, and realistic. AuthBench contains \rededit{428,150 documents} written by \rededit{153,825 individuals} across ten widely used languages, \rededit{9 primary genres}, \rededit{66 fine-grained genres}, and four document-length buckets. It supports two complementary tasks: \emph{authorship attribution}, formulated as same-author retrieval and \emph{authorship verification}, formulated as same-author binary decision. \rededit{We benchmark 47 neural models and three non-neural baselines under a unified zero-shot protocol. Results show that authorship representation remains far from solved: the best retrieval model reaches only 0.258 Success@5, while the best verification model achieves 0.076 EER and 0.968 ROC-AUC. The leaderboard also reveals a meaningful task split, with different model families leading retrieval and verification, and large performance differences across languages, genres, and lengths.} These findings position AuthBench not only as a new benchmark, but as a diagnostic resource for studying when and why authorship representations succeed or fail. We release AuthBench, its evaluation toolkit, and benchmark data at
\url{https://github.com/mao-code/AuthBench} and
\url{https://huggingface.co/datasets/MaoXun/AuthBench}.

\end{abstract}

\section{Introduction}
Authorship representation \citep{huang2025authorship,habib2025trends,tyo2022state} examines whether a model can encode author-specific linguistic and stylistic features into reusable text representations, which is an important capability of current state-of-the-art (SOTA) models. This ability has a wide range of applications, including cybersecurity and digital forensics \citep{abbasi2008writeprints,narayanan2012feasibility,kumar2017army,caliskan2015deanonymizing}, and underpins additional tasks such as plagiarism detection \citep{potthast2014panplagiarism,barron-cedeno-etal-2013-plagiarism} and machine-generated text detection \citep{przystalski2025stylometry,aityan2025lightweight,alshaibani2026arabic,martinek2024deepfakeads}.

Driven by significant progress in large language models (LLMs) \cite{yang2025qwen3,zhang2025qwen3,grattafiori2024llama}, SOTA models have become increasingly generalizable across downstream tasks, languages, and varying input lengths. \rev{However, existing resources still broaden authorship evaluation along only one or two axes at a time. Table~\ref{tab:bench-comparison} summarizes representative resources: classic corpora such as blogs and email collections are often monolingual and domain-specific; PAN established standardized evaluation protocols, but as a family of yearly shared tasks, each edition usually focuses on a narrow task or discourse setting; and more recent resources typically expand coverage along a single dimension, such as multilinguality, cross-discourse evaluation, cross-genre journalism, or document-length control. Consequently, prior work still lacks a unified benchmark that simultaneously offers multilingual coverage, broad genre diversity, explicit length control, and standardized support for both large-scale attribution and verification.}

\begin{table}[t]
\centering
\scriptsize
\setlength{\tabcolsep}{3pt}
\renewcommand{\arraystretch}{1.12}
\resizebox{\linewidth}{!}{%
\begin{tabular}{@{}L{4.0cm}c rr L{2.9cm}cccccc@{}}
\toprule
\textbf{Benchmark} & \textbf{Year} & \textbf{Authors} & \textbf{Docs} & \textbf{Scope} & \multicolumn{3}{c}{\textbf{Coverage}} & \multicolumn{3}{c}{\textbf{Evaluation}} \\
\cmidrule(lr){6-8}\cmidrule(lr){9-11}
 & & & & & \shortstack{\textbf{Multi-}\\\textbf{lingual}} & \shortstack{\textbf{Multi-}\\\textbf{genre}} & \shortstack{\textbf{Length}\\\textbf{bins}} & \shortstack{\textbf{Standard}\\\textbf{splits}} & \textbf{Attribution} & \textbf{Verification} \\
\midrule
\textbf{Blogs50} \citep{zhang-etal-2018-syntax} & 2006 & 50 & 66K & English blogs & \xmark & \xmark & \xmark & \xmark & \cmark & \xmark \\
\textbf{MUD} (adapted corpus) \citep{khan2021deepmetriclearningapproach,riverasoto2021luar} & 2021 & 1.07M & 321.7M & English Reddit & \xmark & \xmark & \xmark & \xmark & \xmark & \xmark \\
\textbf{MARC} \citep{keung2020marc} & 2020 & 1.15M & 1.26M & Multilingual reviews & \cmark & \xmark & \xmark & \cmark & \xmark & \xmark \\
\textbf{PAN 2021 AV} \citep{pan2021av} & 2021 & 278.2K & 73.0K pairs & English fanfiction & \xmark & \xmark & \xmark & \cmark & \xmark & \cmark \\
\textbf{PAN 2022 AV} \citep{pan2022av} & 2022 & 112 & 22.7K pairs & English cross-discourse & \xmark & \cmark & \xmark & \cmark & \xmark & \cmark \\
\textbf{PAN 2023 AV} \citep{pan2023av} & 2023 & 112 & 18.5K pairs & English written/spoken & \xmark & \cmark & \xmark & \cmark & \xmark & \cmark \\
\textbf{CROSSNEWS} \citep{ma2025crossnews} & 2025 & 2.26K & 1.15M & English news + social & \xmark & \cmark & \xmark & \cmark & \cmark & \cmark \\
\textbf{IMDb62} \citep{seroussi2014authorship} & 2011 & 62 & 62K & English reviews & \xmark & \xmark & \xmark & \xmark & \cmark & \xmark \\
\textbf{CMCC} \citep{goldstein-stewart-etal-2008-creating} & 2008 & 21 & 756 & English email/essay/blog/chat & \xmark & \cmark & \xmark & \xmark & \cmark & \xmark \\
\textbf{Guardian} \citep{stamatatos2013robustness} & 2013 & 13 & 1.0K & English opinion/book reviews & \xmark & \cmark & \xmark & \xmark & \cmark & \xmark \\
\textbf{Enron Email} (adapted corpus) \citep{klimt2004enron} & 2004 & $\sim$150 & $\sim$500K & English email & \xmark & \xmark & \xmark & \xmark & \xmark & \xmark \\
\textbf{SMAuC} (adapted corpus) \citep{bevendorff2023smauc} & 2023 & 5.66M & 3.36M & Scientific writing & \xmark & \cmark & \cmark & \xmark & \xmark & \xmark \\
\textbf{AIDBench} (identification benchmark) \citep{wen2024aidbench} & 2024 & 3.25K & 51.5K & English papers/email/blogs/reviews & \xmark & \cmark & \xmark & \cmark & \xmark & \xmark \\
\midrule
\textbf{AuthBench (ours)} & 2026 & \rededit{153.8K} & \rededit{428.2K} & Social, news, literature, blogs, etc. & \cmark & \cmark & \cmark & \cmark & \cmark & \cmark \\
\bottomrule
\end{tabular}
}
\caption{Comparison of AuthBench with representative prior authorship resources.}
\label{tab:bench-comparison}
\end{table}

To address these limitations, we introduce AuthBench, a large-scale multilingual benchmark for authorship representation that supports attribution and verification across genres and document lengths. AuthBench combines author-linked text gathered through our own public-web crawling pipelines with carefully refined portions of existing datasets and benchmark resources, all standardized into a shared schema (see Appendix~\ref{sec:appendix-raw-sources}). After quality filtering, deduplication, language auditing, and balanced sampling, the current release contains \rededit{428,150} documents of diverse lengths (short, medium, long, and extra-long), authored by \rededit{153,825} individuals across ten widely used languages (en, zh, hi, es, fr, ar, ru, de, ja, ko), spanning \rededit{nine primary genres} with \rededit{66 fine-grained sub-genres}. AuthBench supports two complementary evaluation tasks: (i) \emph{authorship attribution}, formulated as \rev{an open-world same-author} retrieval task evaluated with ranking-based metrics (Success@K, Recall@K, nDCG@K, \rededit{and MRR}); and (ii) \emph{authorship verification}, formulated as binary decision over query--candidate pairs and evaluated \rev{with EER and ROC-AUC}. We release standardized dataset splits and a unified evaluation toolkit to facilitate reproducible and comprehensive evaluation.

\rededit{We conduct a comprehensive zero-shot evaluation of 47 neural models and three non-neural baselines on AuthBench, spanning embedding models, base LLMs, and instruction-tuned variants. The leaderboard reveals a clear task split: \texttt{multilingual-e5-large} achieves the best overall retrieval performance (0.258 S@5, 0.254 R@5, 0.217 nDCG@5, 0.220 MRR), while \texttt{llama3.1-8b-instruct} achieves the best overall verification performance (0.076 EER, 0.968 ROC-AUC). Moreover, model quality still varies substantially across languages, document lengths, and genres, highlighting persistent challenges for robust authorship modeling.}

\section{Related Work}
\label{sec:related-work}
\textbf{Authorship representation.} 
Authorship representation \cite{huang2025authorship,habib2025trends,tyo2022state} seeks to identify persistent author-specific signals in written documents. The field is commonly structured around two core evaluation tasks: (i) authorship attribution, which links a document to its author \rev{and is traditionally studied as closed-set classification \citep{stamatatos2009survey,neal2017surveying} but is increasingly instantiated in representation-learning settings as open-world same-author retrieval over a candidate pool \citep{riverasoto2021luar,wang2023canrep}}; and (ii) authorship verification, which determines whether two documents were written by the same author \citep{neal2017surveying,pan2022av}. In parallel, authorship representation learning develops document embeddings that encode stylistic signals and can be used to support both attribution and verification at scale. \rev{Recent work therefore commonly reports retrieval metrics such as Recall@K or MRR for retrieval-style attribution, while standardized verification benchmarks emphasize ROC-based metrics such as AUC and representation-oriented analyses also report EER as a compact summary of the false-accept / false-reject trade-off \citep{pan2022av,pan2023av,wang2023canrep}.}

\textbf{Authorship Benchmark Development.}
\rev{Early authorship resources were instrumental for the field, but they were typically small-scale, monolingual, and tied to a narrow domain \citep{stamatatos2009survey, neal2017surveying}. Classic datasets such as blog and email corpora \citep{schler2006effects,klimt2004enron} enabled important methodological progress, yet they offered limited coverage in language diversity, genre breadth, and explicit length control, and many were not released with broadly adopted benchmark splits. A second challenge is fragmentation: several resources widely used in modern work, including Blogs50, MUD, Enron, and SMAuC, are adapted from broader source collections rather than introduced as unified authorship benchmarks. PAN later established influential shared-task protocols for authorship analysis \citep{potthast2016whowrote,bevendorff2020pan}, but each edition typically focuses on a single task and discourse setting. More recent resources broaden the space only partially: MARC adds multilingual reviews \citep{keung2020marc}, PAN 2022 and PAN 2023 study controlled cross-discourse verification in English \citep{pan2022av,pan2023av}, CROSSNEWS targets cross-genre journalism \citep{ma2025crossnews}, SMAuC emphasizes scientific writing with length control \citep{bevendorff2023smauc}, and AIDBench centers on identification-style evaluations \citep{wen2024aidbench}. What is still missing is a single benchmark that jointly supports multilingual, multi-genre, length-aware, standardized evaluation for both retrieval-style attribution and verification. AuthBench is designed to fill that gap.}

\section{AuthBench}
\label{sec:overview}
In this section, we first describe the AuthBench construction in Section \ref{sec:construction}. Then we present the statistics of the AuthBench in Section \ref{sec:statbench}.
\subsection{Benchmark Construction}
\label{sec:construction}
\begin{figure*}[h!]
  \centering
  \safeincludegraphics[width=\textwidth]{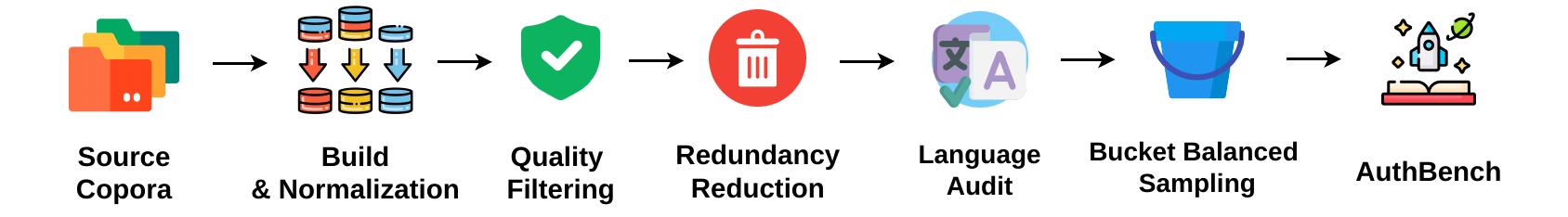}
  \caption{AuthBench construction pipeline. Full details are provided in Appendix~\ref{sec:appendix-construction-details}.}
  \label{fig:processing_flow}
\end{figure*}

\rev{Figure~\ref{fig:processing_flow} summarizes the AuthBench construction workflow, and Figure~\ref{fig:json-examples} gives a quick view of the final record format. AuthBench is built from two complementary inputs: author-linked text gathered through our public-web crawling pipelines, and selected existing datasets or benchmark resources that add useful languages, genres, or length regimes (Appendix~\ref{sec:appendix-raw-sources}). We then standardize everything through five stages: Build \& Normalization, Quality Filtering, Redundancy Reduction, Language Audit, and Bucket Balanced Sampling. Stage 1 maps every source into a shared schema, chunks overly long documents, and caps per-author document counts so heterogeneous materials become comparable without letting prolific authors dominate the benchmark.}

\rededit{The later stages turn that broad collection into a reliable evaluation benchmark. Quality Filtering removes noisy, low-information, and script-mismatched text; Redundancy Reduction removes exact duplicates, near-text overlaps, and optional near-author overlaps to reduce leakage; and the Language Audit verifies or retags suspicious labels. Bucket Balanced Sampling then applies hierarchical targets over language, genre, and length bucket before writing stratified train/dev/test splits, keeping the final release diverse, balanced, and reproducible. In short, the pipeline converts both crawled and inherited resources into a unified benchmark optimized for consistent authorship evaluation. Complete implementation-aligned details are provided in Appendix~\ref{sec:appendix-construction-details}.}

\subsection{Statistics of AuthBench}
\label{sec:statbench}

After construction, AuthBench comprises \rededit{428{,}150 documents authored by 153{,}825 individuals} across 10 languages: English (en), Spanish (es), Chinese (zh), French (fr), German (de), Arabic (ar), Russian (ru), Japanese (ja), Korean (ko), and Hindi (hi). AuthBench spans \rededit{9 primary genres}, including \rededit{\texttt{social\_media}, \texttt{literature}, \texttt{news}, \texttt{blog}, \texttt{media\_reviews}, \texttt{poetry}, \texttt{ecommerce\_reviews}, \texttt{qna}, and \texttt{research\_paper}}. \rededit{The distribution is dominated by \texttt{social\_media} (40.9\%), \texttt{literature} (30.0\%), and \texttt{news} (17.2\%), with the remaining six primary genres accounting for the final 11.9\% of documents.} Documents in AuthBench are further categorized into four token-length buckets: short (1–10 tokens), medium (11–100), long (101–500), and extra-long ($>$500). An overview of AuthBench statistics is presented in Table~\ref{tab:aurabench-overview}. \rededit{The current split materialization contains 198{,}345 query documents and 229{,}805 candidate documents, partitioned into train/dev/test as 156{,}335/21{,}008/21{,}002 queries and 186{,}184/21{,}813/21{,}808 candidates.} Additional detailed statistics are provided in Appendix~\ref{sec:appendix-additional-stats}.

\begin{table}[t]
  \centering
  \setlength{\tabcolsep}{6pt}
  \begin{tabular}{lrr}
    \toprule
    \textbf{Category} & \textbf{\#Docs} & \textbf{Share} \\
    \midrule
    \multicolumn{3}{l}{\textit{Overall}} \\
    Total & \rededit{428{,}150} & 100.0\% \\
    Languages & 10 & -- \\
    Authors & \rededit{153{,}825} & -- \\
    Avg.\ docs per author & \rededit{2.78} & -- \\
    Genres (fine-grained) & \rededit{66} & -- \\
    \midrule
    \multicolumn{3}{l}{\textit{By language}} \\
    English (en) & \rededit{97{,}974} & \rededit{22.9\%} \\
    Spanish (es) & \rededit{33{,}395} & \rededit{7.8\%} \\
    Chinese (zh) & \rededit{55{,}368} & \rededit{12.9\%} \\
    French (fr) & \rededit{31{,}225} & \rededit{7.3\%} \\
    German (de) & \rededit{39{,}813} & \rededit{9.3\%} \\
    Arabic (ar) & \rededit{42{,}091} & \rededit{9.8\%} \\
    Russian (ru) & \rededit{66{,}084} & \rededit{15.4\%} \\
    Japanese (ja) & \rededit{21{,}494} & \rededit{5.0\%} \\
    Korean (ko) & \rededit{33{,}881} & \rededit{7.9\%} \\
    Hindi (hi) & \rededit{6{,}825} & \rededit{1.6\%} \\
    \midrule
    \multicolumn{3}{l}{\textit{By length}} \\
    Short & \rededit{24{,}476} & \rededit{5.7\%} \\
    Medium & \rededit{211{,}910} & \rededit{49.5\%} \\
    Long & \rededit{184{,}447} & \rededit{43.1\%} \\
    Extra-long & \rededit{7{,}317} & \rededit{1.7\%} \\
    \bottomrule
  \end{tabular}
\caption{Statistics of our AuthBench.}
  \label{tab:aurabench-overview}
\end{table}

\begin{figure*}[t]
  \centering
  \safeincludegraphics[width=\textwidth]{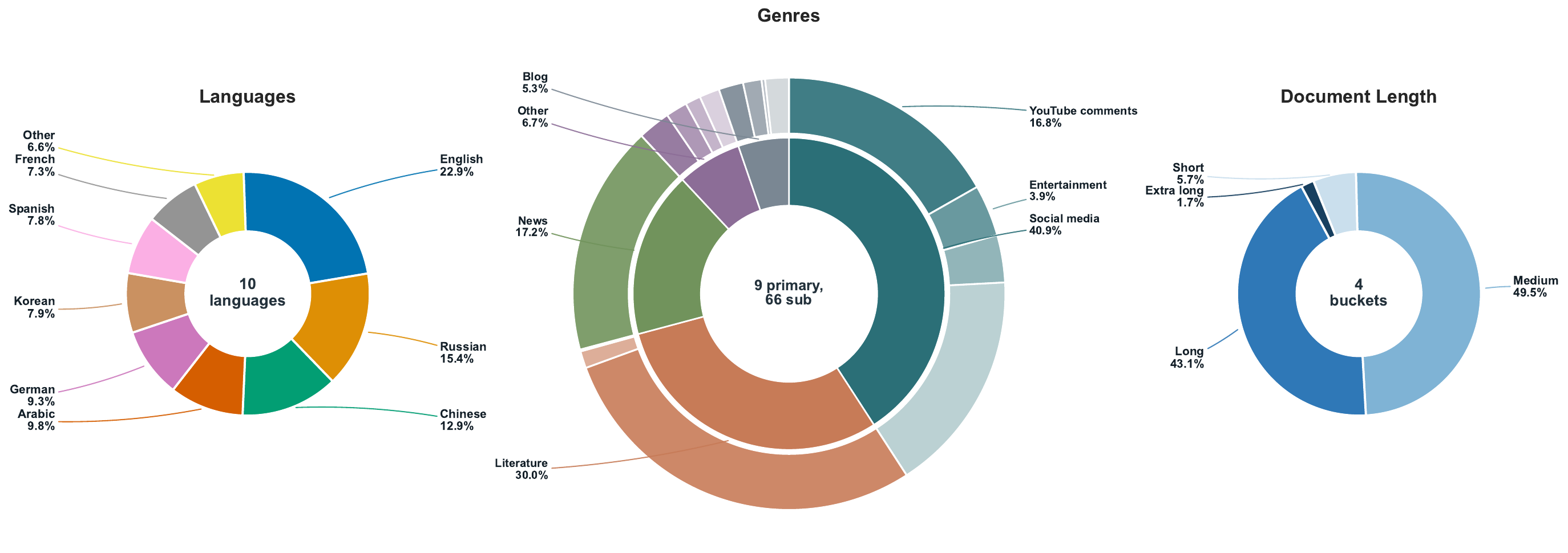}
  \caption{Overview of the current AuthBench profile for document proportion across languages, genres and document lengths.}
  \label{fig:benchmark-profile-overview}
\end{figure*}

Figure~\ref{fig:benchmark-profile-overview} complements Table~\ref{tab:aurabench-overview} with a compact visual summary of the benchmark profile. The language-level counts show that AuthBench remains broadly distributed across all ten languages rather than collapsing into a single dominant language, while the language--genre pie-chart reveals clear source-driven heterogeneity in genre coverage across languages. The token-length plot further shows that most documents concentrate in the medium-to-long range, with shorter texts and extra-long documents appearing less frequently. Together, these views highlight that AuthBench is broad in coverage but intentionally preserves realistic distributional variation across languages, genres, and lengths.

\subsection{Pipeline Summary}



\begin{figure}[h!]
\centering

\begin{minipage}[t]{0.48\linewidth}
\begin{jsonexample}{}
\begin{JSONVerbatim}
{
  "candidate_id": "doc_116326",
  "author_id": "43b5e789...c39320",
  "lang": "en",
  "genre": "social_media/people",
  "content": "How recently are these things you found? People have pasts ... How would you like it?",
  "source": "exorde",
  "token_length": 156
}
\end{JSONVerbatim}
\end{jsonexample}
\end{minipage}\hfill
\begin{minipage}[t]{0.48\linewidth}
\begin{jsonexample}{}
\begin{JSONVerbatim}
{
  "candidate_id": "doc_274629",
  "author_id": "c423340f...30add",
  "lang": "zh",
  "genre": "news",
  "content": "由于中国政府加大对游戏内容的管控力度，苹果公司正将数千款游戏应用从其中国平台上下架...",
  "source": "babel_briefings",
  "token_length": 74
}
\end{JSONVerbatim}
\end{jsonexample}
\end{minipage}

\caption{Examples in our AuthBench.}
\label{fig:json-examples}
\end{figure}

\section{Experiment Setup}
\textbf{Evaluation details.}  We embed queries and candidates with each model’s encoder (or the final hidden states of LLM backbones) using a unified pipeline: tokenize each input, extract the last hidden states, pool into a single vector (mean pooling by default), and L2-normalize the resulting embedding. Similarities are computed with \rev{cosine similarity} between normalized embeddings. \rev{We use full-sized within-split candidate pool; verification metrics are computed over all same-author and non-matching query--candidate pairs in that pool. The complete experiment setup is defined in Appendix~\ref{subsec:eval-protocol}.}

\noindent\textbf{Models.} \rededit{We evaluate 47 neural models together with three non-neural baselines: a character $3$--$5$ gram TF-IDF cosine baseline \citep{salton1988termweighting}, a lightweight stylometric n-gram baseline inspired by \citet{koppel2004authorship}, and a PPM-style compression baseline following \citet{teahan2003compression}. These systems complement the neural leaderboard with strong lexical and compression-based reference points.} Full model details are provided in Table~\ref{tab:models-evaluated} of Appendix~\ref{subsec:eval-protocol}.

\noindent\textbf{Metrics.} AuthBench supports both authorship attribution and authorship verification tasks. 
For authorship attribution, \rededit{we report Success@5 (S@5), Recall@5 (R@5), nDCG@5, and MRR;} \rev{these respectively capture shortlist utility, coverage of multiple same-author targets, ranking quality near the top of the list, and the rank of the first correct same-author hit.}
For authorship verification, \rev{we compute EER and ROC-AUC; EER summarizes the balanced operating point where false acceptance and false rejection are equal, while ROC-AUC measures threshold-independent separability between same-author and different-author pairs.}
\rededit{The main leaderboard reports all six aggregate metrics, while the slice tables below focus on S@5 for compactness.} \rev{Formal definitions of all evaluation metrics are provided in Appendix~\ref{subsec:metrics}.}

\section{Experimental Results}
\label{sec:experiments}
\rededit{All results in this section are reported on the AuthBench test split, which contains 21,002 queries and 21,808 candidates.} In this section, we first present the overall performance of SOTA models on AuthBench in Section~\ref{sec:mainresults}. We then provide a detailed analysis of performance broken down by language, genre, and document length. Finally, in Section~\ref{sec:dynamics}, \rededit{we clarify the scope of the current zero-shot study and outline post-training as future work rather than as a reported main result.}

\begin{table}[t]
\centering
\small
\setlength{\tabcolsep}{5pt}
{\color{editred}
\resizebox{\linewidth}{!}{%
\begin{tabular}{llrrrrrr}
\toprule
\textbf{Model} & \textbf{Model Size} & \textbf{S@5 $\uparrow$} & \textbf{R@5 $\uparrow$} & \textbf{nDCG@5 $\uparrow$} & \textbf{MRR $\uparrow$} & \textbf{ROC-AUC $\uparrow$} & \textbf{EER $\downarrow$} \\
\midrule
\multicolumn{8}{l}{\textit{LLMs (instruction-tuned)}} \\
\midrule
\texttt{llama3-8b-instruct} & 8B & \textbf{0.251} & \textbf{0.247} & \textbf{0.206} & \textbf{0.209} & \underline{0.967} & \underline{0.080} \\
\texttt{llama3.1-8b-instruct} & 8B & \underline{0.247} & \underline{0.243} & \underline{0.204} & \underline{0.208} & \textbf{0.968} & \textbf{0.076} \\
\texttt{qwen2.5-7b-instruct} & 7.6B & 0.207 & 0.203 & 0.168 & 0.172 & 0.964 & 0.083 \\
\texttt{qwen3-4b-instruct} & 4B & 0.197 & 0.193 & 0.159 & 0.163 & 0.958 & 0.094 \\
\midrule
\multicolumn{8}{l}{\textit{LLMs (base)}} \\
\midrule
\texttt{llama3-8b} & 8B & \textbf{0.254} & \textbf{0.250} & \textbf{0.210} & \textbf{0.213} & \textbf{0.967} & \textbf{0.079} \\
\texttt{llama3.1-8b} & 8B & \underline{0.253} & \underline{0.249} & \underline{0.209} & \underline{0.212} & \underline{0.967} & \underline{0.080} \\
\texttt{deepseek-llm-7b-base} & 7B & 0.220 & 0.216 & 0.180 & 0.184 & 0.954 & 0.105 \\
\texttt{qwen3-4b} & 4B & 0.210 & 0.206 & 0.171 & 0.175 & 0.960 & 0.087 \\
\midrule
\multicolumn{8}{l}{\textit{Embedding models (instruction-tuned)}} \\
\midrule
\texttt{e5-mistral-7b-instruct} & 7.1B & \textbf{0.242} & \textbf{0.238} & \textbf{0.202} & \textbf{0.205} & \underline{0.955} & \underline{0.096} \\
\texttt{gte-qwen2-7b-instruct} & 7.6B & \underline{0.240} & \underline{0.235} & \underline{0.198} & \underline{0.202} & \textbf{0.961} & \textbf{0.078} \\
\midrule
\multicolumn{8}{l}{\textit{Embedding models}} \\
\midrule
\texttt{multilingual-e5-large} & 559.9M & \textbf{0.258} & \textbf{0.254} & \textbf{0.217} & \textbf{0.220} & 0.920 & 0.157 \\
\texttt{multilingual-e5-base} & 278M & \underline{0.250} & \underline{0.245} & \underline{0.209} & \underline{0.212} & 0.918 & 0.161 \\
\texttt{qwen3-embedding-8b} & 7.6B & 0.241 & 0.237 & 0.201 & 0.203 & \underline{0.940} & \underline{0.135} \\
\texttt{sfr-embedding-mistral} & 7.1B & 0.240 & 0.236 & 0.201 & 0.205 & \textbf{0.956} & \textbf{0.096} \\
\midrule
\multicolumn{8}{l}{\textit{Lexical / non-neural baselines}} \\
\midrule
\texttt{tfidf} & -- & \textbf{0.197} & \textbf{0.194} & \textbf{0.167} & \textbf{0.183} & \underline{0.837} & \underline{0.219} \\
\texttt{ngram} & -- & \underline{0.170} & \underline{0.167} & \underline{0.145} & \underline{0.149} & \textbf{0.874} & \textbf{0.213} \\
\texttt{ppm} & -- & 0.157 & 0.156 & 0.137 & 0.142 & 0.792 & 0.298 \\
\bottomrule
\end{tabular}
}
}
\caption{\color{editred}Main results on AuthBench (test split). The retrieval leaderboard now reports S@5, R@5, nDCG@5, and MRR; the verification leaderboard reports both ROC-AUC and EER. \textbf{Bold} entries mark the best result within each model group for a metric, and \underline{underlined} entries mark the second best. The full 50-model evaluation, including all three non-neural baselines, is deferred to Appendix~\ref{sec:appendix-full-results}.}
\label{tab:overall-leaderboard}
\end{table}

\subsection{Main Results}
\label{sec:mainresults}
\rededit{\textbf{Finding 1: retrieval and verification favor different model families.}}

\rededit{Table~\ref{tab:overall-leaderboard} shows a clear split between ranking and verification ability. For retrieval, the strongest model is \texttt{multilingual-e5-large}, reaching 0.258 S@5, 0.254 R@5, 0.217 nDCG@5, and 0.220 MRR. The top base LLMs are extremely close: \texttt{llama3-8b} and \texttt{llama3.1-8b} reach 0.254 and 0.253 S@5, respectively. This narrow gap suggests that large generative backbones already encode strong authorship cues, but specialized multilingual embedding models remain slightly better at turning those cues into stable nearest-neighbor structure.}

\rededit{Verification tells a different story. \texttt{llama3.1-8b-instruct} is the strongest verifier with 0.076 EER and 0.968 ROC-AUC, followed closely by \texttt{llama3-8b-instruct} and \texttt{llama3-8b}. In contrast, the best retrieval model is not the best verifier. This is an important takeaway for future work: authorship retrieval and authorship verification should not be treated as interchangeable probes of the same representation quality. Retrieval rewards global neighborhood structure, whereas verification also depends on pairwise calibration and decision-boundary sharpness.}

\rededit{The non-neural baselines remain informative but clearly behind the best neural systems. \texttt{tfidf} is the strongest lexical retrieval baseline at 0.197 S@5, while \texttt{ngram} is the strongest lexical verifier at 0.213 EER and 0.874 ROC-AUC. The gap to the neural frontier is still substantial, indicating that modern encoders capture authorial regularities beyond surface lexical overlap. At the same time, even the best neural scores remain far from ceiling, which suggests that robust authorship representation is still an open problem rather than a solved byproduct of general text embedding quality.}

\begin{table}[t]
\centering
\small
\setlength{\tabcolsep}{4pt}
{\color{editred}
\resizebox{\linewidth}{!}{%
\begin{tabular}{llrrrrrrrrrr}
\toprule
\textbf{Model} & \textbf{Model Size} & \textbf{\texttt{ar}} & \textbf{\texttt{de}} & \textbf{\texttt{en}} & \textbf{\texttt{es}} & \textbf{\texttt{fr}} & \textbf{\texttt{hi}} & \textbf{\texttt{ja}} & \textbf{\texttt{ko}} & \textbf{\texttt{ru}} & \textbf{\texttt{zh}} \\
\midrule
\multicolumn{12}{l}{\textit{LLMs (instruction-tuned)}} \\
\midrule
\texttt{llama3-8b-instruct} & 8B & \textbf{0.226} & \underline{0.244} & \textbf{0.261} & \textbf{0.265} & \textbf{0.269} & \underline{0.289} & \textbf{0.243} & \underline{0.173} & \underline{0.164} & \textbf{0.382} \\
\texttt{llama3.1-8b-instruct} & 8B & \underline{0.211} & \textbf{0.245} & \underline{0.259} & \underline{0.265} & \underline{0.265} & \textbf{0.297} & \underline{0.229} & \textbf{0.177} & \textbf{0.165} & \underline{0.373} \\
\texttt{qwen2.5-7b-instruct} & 7.6B & 0.182 & 0.210 & 0.201 & 0.208 & 0.214 & 0.215 & 0.215 & 0.143 & 0.143 & 0.343 \\
\texttt{qwen3-4b-instruct} & 4B & 0.187 & 0.195 & 0.187 & 0.203 & 0.198 & 0.229 & 0.212 & 0.143 & 0.143 & 0.304 \\
\midrule
\multicolumn{12}{l}{\textit{LLMs (base)}} \\
\midrule
\texttt{deepseek-llm-7b-base} & 7B & 0.183 & 0.216 & 0.238 & 0.223 & 0.220 & 0.238 & 0.200 & 0.144 & 0.131 & \underline{0.367} \\
\texttt{llama3-8b} & 8B & \underline{0.221} & \underline{0.247} & \textbf{0.270} & \textbf{0.272} & \textbf{0.270} & \underline{0.289} & \underline{0.242} & \underline{0.183} & \underline{0.165} & \textbf{0.380} \\
\texttt{llama3.1-8b} & 8B & \textbf{0.225} & \textbf{0.249} & \underline{0.270} & \underline{0.272} & \underline{0.266} & \textbf{0.297} & \textbf{0.246} & \textbf{0.186} & \textbf{0.173} & 0.361 \\
\texttt{qwen3-4b} & 4B & 0.189 & 0.217 & 0.204 & 0.213 & 0.214 & 0.227 & 0.220 & 0.142 & 0.147 & 0.336 \\
\midrule
\multicolumn{12}{l}{\textit{Embedding models (instruction-tuned)}} \\
\midrule
\texttt{e5-mistral-7b-instruct} & 7.1B & \textbf{0.223} & \textbf{0.241} & \textbf{0.237} & \textbf{0.254} & \textbf{0.250} & \textbf{0.272} & \underline{0.223} & \textbf{0.164} & \underline{0.164} & \textbf{0.396} \\
\texttt{gte-qwen2-7b-instruct} & 7.6B & \underline{0.220} & \underline{0.241} & \underline{0.232} & \underline{0.235} & \underline{0.239} & \underline{0.252} & \textbf{0.249} & \underline{0.156} & \textbf{0.171} & \underline{0.396} \\
\midrule
\multicolumn{12}{l}{\textit{Embedding models}} \\
\midrule
\texttt{multilingual-e5-base} & 278M & \underline{0.251} & 0.210 & 0.226 & \underline{0.267} & \underline{0.262} & 0.238 & \underline{0.253} & \underline{0.185} & \underline{0.188} & 0.413 \\
\texttt{multilingual-e5-large} & 559.9M & \textbf{0.252} & 0.221 & \underline{0.231} & \textbf{0.284} & \textbf{0.280} & \textbf{0.263} & \textbf{0.268} & \textbf{0.196} & \textbf{0.190} & \underline{0.422} \\
\texttt{qwen3-embedding-8b} & 7.6B & 0.239 & \textbf{0.242} & 0.209 & 0.239 & 0.241 & 0.244 & 0.251 & 0.169 & 0.170 & \textbf{0.425} \\
\texttt{sfr-embedding-mistral} & 7.1B & 0.221 & \underline{0.236} & \textbf{0.236} & 0.255 & 0.248 & \underline{0.258} & 0.214 & 0.162 & 0.162 & 0.397 \\
\midrule
\multicolumn{12}{l}{\textit{Lexical / non-neural baselines}} \\
\midrule
\texttt{ngram} & -- & \underline{0.238} & 0.137 & 0.126 & 0.162 & \underline{0.166} & 0.244 & \underline{0.160} & \underline{0.168} & \underline{0.132} & \underline{0.267} \\
\texttt{ppm} & -- & 0.229 & \underline{0.140} & \textbf{0.149} & \underline{0.190} & \textbf{0.172} & \underline{0.255} & 0.113 & 0.103 & 0.118 & 0.184 \\
\texttt{tfidf} & -- & \textbf{0.254} & \textbf{0.148} & \underline{0.146} & \textbf{0.190} & 0.155 & \textbf{0.280} & \textbf{0.226} & \textbf{0.184} & \textbf{0.154} & \textbf{0.346} \\
\bottomrule
\end{tabular}
}
}
\caption{\color{editred}Results by language (S@5). The updated slice table keeps a compact representative subset; complete language-wise S@5 and EER tables are reported in Appendix~\ref{sec:appendix-full-results}.}
\label{tab:lang-s10-top}
\end{table}

\subsection{Results by Language}
\rededit{\textbf{Finding 2: performance varies sharply across languages, and the best model differs by language.}}
\rededit{Table~\ref{tab:lang-s10-top} shows that there is no single universally best model across languages. \texttt{multilingual-e5-large} leads on Spanish, French, Japanese, Korean, and Russian; \texttt{qwen3-embedding-8b} is strongest on Chinese; \texttt{llama3-8b} is strongest on English; \texttt{llama3.1-8b} leads on German and Hindi; and the lexical \texttt{tfidf} baseline is narrowly strongest on Arabic. This pattern suggests that multilingual authorship representation is not just a matter of scaling one architecture, but of matching representation bias to the linguistic and genre composition of each language slice.}

\rededit{The spread in difficulty is also large. Chinese is comparatively easy, with \texttt{qwen3-embedding-8b} reaching 0.425 S@5, whereas Russian and Korean remain difficult even for the best systems, topping out at 0.190 and 0.196. These differences imply that benchmark-wide averages can hide substantial cross-lingual brittleness. Future research should therefore treat multilingual authorship modeling as a robustness problem, not simply as macro-averaged multilingual transfer.}

\rededit{A second notable trend is that multilingual embedding models dominate much of the non-English landscape. Their advantage over LLMs is especially visible in Chinese, Spanish, French, Japanese, Korean, and Russian. This suggests that explicit multilingual representation alignment remains highly valuable for authorship retrieval, and that future progress may come from improving language-sensitive embedding geometry rather than relying on larger generative models alone.}

\begin{table}[t]
\centering
\small
\setlength{\tabcolsep}{4pt}
{\color{editred}
\resizebox{\linewidth}{!}{%
\begin{tabular}{llrrrrrrrrr}
\toprule
\textbf{Model} & \textbf{Model Size} & \textbf{\texttt{blog}} & \textbf{\texttt{ecommerce\_reviews}} & \textbf{\texttt{literature}} & \textbf{\texttt{media\_reviews}} & \textbf{\texttt{news}} & \textbf{\texttt{poetry}} & \textbf{\texttt{qna}} & \textbf{\texttt{research\_paper}} & \textbf{\texttt{social\_media}} \\
\midrule
\multicolumn{11}{l}{\textit{LLMs (instruction-tuned)}} \\
\midrule
\texttt{llama3-8b-instruct} & 8B & \underline{0.213} & \underline{0.107} & \textbf{0.521} & \textbf{0.070} & \textbf{0.302} & \textbf{0.639} & \textbf{0.650} & \underline{0.769} & \textbf{0.206} \\
\texttt{llama3.1-8b-instruct} & 8B & \textbf{0.220} & \textbf{0.112} & \underline{0.518} & 0.035 & \underline{0.293} & 0.583 & \underline{0.618} & \textbf{0.787} & \underline{0.204} \\
\texttt{qwen2.5-7b-instruct} & 7.6B & 0.162 & 0.084 & 0.496 & 0.035 & 0.242 & 0.583 & 0.567 & 0.719 & 0.167 \\
\texttt{qwen3-4b-instruct} & 4B & 0.149 & 0.080 & 0.462 & \underline{0.053} & 0.227 & \underline{0.611} & 0.548 & 0.688 & 0.160 \\
\midrule
\multicolumn{11}{l}{\textit{LLMs (base)}} \\
\midrule
\texttt{deepseek-llm-7b-base} & 7B & 0.215 & \underline{0.114} & 0.486 & 0.035 & 0.257 & 0.639 & 0.548 & 0.719 & 0.176 \\
\texttt{llama3-8b} & 8B & \underline{0.229} & \textbf{0.119} & \underline{0.524} & \textbf{0.070} & \textbf{0.311} & \textbf{0.667} & \textbf{0.631} & \underline{0.800} & \textbf{0.207} \\
\texttt{llama3.1-8b} & 8B & \textbf{0.232} & 0.112 & \textbf{0.532} & \underline{0.053} & \underline{0.305} & \underline{0.667} & \underline{0.631} & \textbf{0.812} & \underline{0.207} \\
\texttt{qwen3-4b} & 4B & 0.166 & 0.084 & 0.485 & 0.035 & 0.241 & 0.639 & 0.561 & 0.738 & 0.171 \\
\midrule
\multicolumn{11}{l}{\textit{Embedding models (instruction-tuned)}} \\
\midrule
\texttt{e5-mistral-7b-instruct} & 7.1B & \textbf{0.202} & \textbf{0.110} & \underline{0.514} & \textbf{0.053} & \underline{0.262} & \underline{0.583} & \underline{0.580} & \underline{0.725} & \textbf{0.207} \\
\texttt{gte-qwen2-7b-instruct} & 7.6B & \underline{0.191} & \underline{0.091} & \textbf{0.538} & \underline{0.035} & \textbf{0.281} & \textbf{0.611} & \textbf{0.586} & \textbf{0.781} & \underline{0.197} \\
\midrule
\multicolumn{11}{l}{\textit{Embedding models}} \\
\midrule
\texttt{multilingual-e5-base} & 278M & 0.146 & \textbf{0.107} & 0.453 & \underline{0.053} & \underline{0.258} & 0.500 & \underline{0.631} & 0.756 & \underline{0.230} \\
\texttt{multilingual-e5-large} & 559.9M & \underline{0.160} & 0.100 & 0.479 & 0.053 & \textbf{0.268} & 0.444 & \textbf{0.682} & \underline{0.769} & \textbf{0.236} \\
\texttt{qwen3-embedding-8b} & 7.6B & 0.151 & 0.091 & \textbf{0.536} & \textbf{0.088} & 0.256 & \underline{0.556} & 0.586 & \textbf{0.838} & 0.209 \\
\texttt{sfr-embedding-mistral} & 7.1B & \textbf{0.199} & \underline{0.105} & \underline{0.512} & 0.053 & 0.257 & \textbf{0.583} & 0.567 & 0.756 & 0.207 \\
\midrule
\multicolumn{11}{l}{\textit{Lexical / non-neural baselines}} \\
\midrule
\texttt{ngram} & -- & 0.066 & 0.050 & \underline{0.289} & \underline{0.018} & \underline{0.161} & \underline{0.278} & 0.325 & 0.431 & \underline{0.168} \\
\texttt{ppm} & -- & \textbf{0.090} & \textbf{0.064} & 0.230 & 0.000 & 0.149 & 0.250 & \underline{0.420} & \textbf{0.631} & 0.153 \\
\texttt{tfidf} & -- & \underline{0.089} & \underline{0.053} & \textbf{0.323} & \textbf{0.035} & \textbf{0.169} & \textbf{0.444} & \textbf{0.459} & \underline{0.613} & \textbf{0.198} \\
\bottomrule
\end{tabular}
}
}
\caption{\color{editred}Results by primary genre (S@5). Complete genre-wise S@5 and EER tables are reported in Appendix~\ref{sec:appendix-full-results}.}
\label{tab:genre-s10-top}
\end{table}

\begin{table}[t]
\centering
\small
\setlength{\tabcolsep}{4pt}
{\color{editred}
\begin{tabular}{llrrrr}
\toprule
\textbf{Model} & \textbf{Model Size} & \textbf{\texttt{short}} & \textbf{\texttt{medium}} & \textbf{\texttt{long}} & \textbf{\texttt{extra\_long}} \\
\midrule
\multicolumn{6}{l}{\textit{LLMs (instruction-tuned)}} \\
\midrule
\texttt{llama3-8b-instruct} & 8B & \textbf{0.150} & \textbf{0.199} & \textbf{0.432} & \textbf{0.539} \\
\texttt{llama3.1-8b-instruct} & 8B & \underline{0.142} & \underline{0.196} & \underline{0.427} & \underline{0.531} \\
\texttt{qwen2.5-7b-instruct} & 7.6B & 0.103 & 0.159 & 0.379 & 0.496 \\
\texttt{qwen3-4b-instruct} & 4B & 0.081 & 0.150 & 0.368 & 0.488 \\
\midrule
\multicolumn{6}{l}{\textit{LLMs (base)}} \\
\midrule
\texttt{deepseek-llm-7b-base} & 7B & 0.121 & 0.169 & 0.393 & \underline{0.528} \\
\texttt{llama3-8b} & 8B & \textbf{0.150} & \textbf{0.202} & \underline{0.436} & 0.524 \\
\texttt{llama3.1-8b} & 8B & \underline{0.135} & \underline{0.202} & \textbf{0.439} & \textbf{0.543} \\
\texttt{qwen3-4b} & 4B & 0.101 & 0.161 & 0.385 & 0.496 \\
\midrule
\multicolumn{6}{l}{\textit{Embedding models (instruction-tuned)}} \\
\midrule
\texttt{e5-mistral-7b-instruct} & 7.1B & \textbf{0.161} & \textbf{0.196} & \underline{0.398} & \textbf{0.524} \\
\texttt{gte-qwen2-7b-instruct} & 7.6B & \underline{0.137} & \underline{0.192} & \textbf{0.409} & \underline{0.500} \\
\midrule
\multicolumn{6}{l}{\textit{Embedding models}} \\
\midrule
\texttt{multilingual-e5-base} & 278M & \textbf{0.187} & \underline{0.213} & 0.376 & 0.417 \\
\texttt{multilingual-e5-large} & 559.9M & \underline{0.182} & \textbf{0.220} & 0.392 & 0.476 \\
\texttt{qwen3-embedding-8b} & 7.6B & 0.157 & 0.197 & \textbf{0.395} & \underline{0.504} \\
\texttt{sfr-embedding-mistral} & 7.1B & 0.168 & 0.193 & \underline{0.395} & \textbf{0.520} \\
\midrule
\multicolumn{6}{l}{\textit{Lexical / non-neural baselines}} \\
\midrule
\texttt{ngram} & -- & \underline{0.151} & \underline{0.136} & \underline{0.271} & \underline{0.331} \\
\texttt{ppm} & -- & 0.079 & 0.134 & 0.255 & 0.240 \\
\texttt{tfidf} & -- & \textbf{0.157} & \textbf{0.164} & \textbf{0.300} & \textbf{0.437} \\
\bottomrule
\end{tabular}
}
\caption{\color{editred}Results by length bucket (S@5). Complete length-wise S@5 and EER tables are reported in Appendix~\ref{sec:appendix-full-results}.}
\label{tab:length-s10-top}
\end{table}

\subsection{Results by Genre}
\rededit{\textbf{Finding 3: genre difficulty varies dramatically, and different genres favor different architectures.}}
\rededit{Table~\ref{tab:genre-s10-top} reveals large genre-dependent differences in difficulty. The easiest slice is \texttt{research\_paper}, where \texttt{qwen3-embedding-8b} reaches 0.838 S@5, followed by \texttt{qna} at 0.682 and \texttt{poetry} at 0.667. At the other extreme, \texttt{media\_reviews} and \texttt{ecommerce\_reviews} are much harder, with best S@5 values of only 0.088 and 0.119. This gap indicates that authorship signals are much easier to recover in domains with stronger personal regularity or more stable discourse conventions than in short, noisy, or highly template-driven review settings.}

\rededit{The identity of the best model also changes sharply by genre. \texttt{gte-qwen2-7b-instruct} is strongest on \texttt{literature}; \texttt{qwen3-embedding-8b} is strongest on \texttt{media\_reviews} and \texttt{research\_paper}; \texttt{multilingual-e5-large} leads on \texttt{qna} and \texttt{social\_media}; and the \texttt{llama3} family leads \texttt{blog}, \texttt{news}, \texttt{poetry}, and \texttt{ecommerce\_reviews}. Rather than pointing to one dominant architecture, the table suggests that different model families are capturing different kinds of authorial evidence.}

\rededit{A useful way to interpret this pattern is through discourse structure. Embedding models appear especially strong in structured, information-dense settings such as \texttt{research\_paper} and \texttt{qna}, where topical organization and repeated compositional habits may be easier to preserve in fixed-vector spaces. LLMs remain very competitive in more stylistically expressive genres such as \texttt{blog}, \texttt{news}, and \texttt{poetry}, where broader contextual modeling may better preserve subtle stylistic signatures. Future authorship benchmarks and methods should therefore pay closer attention to genre as a first-class modeling variable rather than a secondary reporting slice.}

\subsection{Performance by Length}
\rededit{\textbf{Finding 4: longer documents are markedly easier than short ones, but the best model still depends on length regime.}} Table~\ref{tab:length-s10-top} presents the updated S@5 breakdown across four document lengths. \rededit{The strongest short-document model is \texttt{multilingual-e5-base} at 0.187 S@5, the strongest medium-document model is \texttt{multilingual-e5-large} at 0.220, and the strongest long and extra-long models are \texttt{llama3.1-8b} at 0.439 and 0.543, respectively.}

\rededit{The length trend is strong and intuitive: authorship retrieval becomes much easier as more text is available. Extra-long documents are the easiest regime overall, and the gain is especially pronounced for the LLMs. For example, \texttt{llama3.1-8b} rises from 0.135 S@5 on short documents to 0.543 on extra-long ones. This suggests that much of the remaining challenge in authorship modeling comes from sparse-evidence settings, where models must identify stable stylistic cues from very limited text.}

\rededit{The strongest model also changes with length. Short and medium documents favor embedding models, with \texttt{multilingual-e5-base} and \texttt{multilingual-e5-large} leading those buckets, whereas long and extra-long documents favor \texttt{llama3.1-8b}. This suggests a useful modeling hypothesis for future work: compact embedding models may be better at extracting robust local stylistic cues when evidence is scarce, while larger LLMs benefit more from longer contexts that expose higher-order discourse and syntactic habits.}

\rededit{Verification follows the same overall pattern. The best EER improves from 0.104 on \texttt{short} to 0.078 on \texttt{medium}, 0.060 on \texttt{long}, and 0.048 on \texttt{extra\_long}. Short documents therefore remain the clearest bottleneck for both retrieval and verification. Advancing this regime will likely require methods explicitly designed for low-evidence authorship signals, rather than simply scaling existing encoders.}

\subsection{\rededit{Post-Training Outlook}}
\label{sec:dynamics}
\rededit{In this paper, AuthBench serves as a large-scale pre-training evaluation benchmark for a broad set of authorship representation models, including embedding models, base LLMs, and instruction-tuned variants. This zero-shot comparison provides a useful foundation for future researchers to choose strong base models under different languages, genres, lengths, and task settings before performing post-training. We are also conducting follow-up analyses on post-training behavior and alternative authorship methods, and we plan to release a second version of the paper with a broader set of post-training results.}

\section{Conclusion}
\rededit{We introduced \textbf{AuthBench}, a large-scale benchmark for evaluating authorship representations across languages, genres, and document lengths under a unified retrieval-and-verification framework. Our zero-shot results show that authorship representation is still far from solved, that retrieval and verification favor different model families, and that robustness across languages, genres, and especially short documents remains the central challenge. We hope AuthBench provides a strong foundation for future work on more reliable and better calibrated authorship modeling in realistic multilingual settings.}

\section*{Limitations} \rededit{A key limitation of AuthBench is that language, genre, and document length are not fully orthogonal axes in the current release. Although we intentionally collected diverse sources and applied a robust filtering, deduplication, and auditing pipeline to reduce avoidable bias, some residual covariance across these axes is difficult to eliminate in practice. As a result, certain slice differences may still reflect source-level correlations rather than authorship difficulty alone. In addition, despite our leakage-reduction pipeline, some residual leakage may remain at scale. We view these as important targets for future benchmark refinement.}


\bibliography{colm2026_conference}
\bibliographystyle{colm2026_conference}

\clearpage
\onecolumn
\appendix

\section{AuthBench Construction Details}
\label{sec:appendix-construction-details}

This appendix provides the full construction specification for AuthBench, including Build \& Normalization, Quality Filtering, Redundancy Reduction, Language Audit, and Bucket Balanced Sampling.
Section~\ref{sec:construction} in the main paper contains a concise summary.

{\color{editred}AuthBench is constructed through a five-stage pipeline: Build \& Normalization, Quality Filtering, Redundancy Reduction, Language Audit, and Bucket Balanced Sampling. The pipeline is designed for large-scale ingestion while applying final benchmark size control only after Quality Filtering and Redundancy Reduction.}

\subsection{Stage 1: Build \& Normalization - Source Ingestion and Normalization}
\label{subsec:construct-ingest}
{\color{editred}Let $\mathcal{S}=\{S_1,\dots,S_m\}$ denote the set of raw sources. Each source yields records with source-specific metadata, but the pipeline converts every surviving item into a unified document representation}
\[
{\color{editred}x=(r,\ a,\ \ell,\ g,\ \sigma,\ t,\ L(t),\ k(t),\ \mathbf{m}),}
\]
{\color{editred}where $r$ is a source-level raw identifier, $a$ is an anonymized author identifier, $\ell$ is language, $g$ is the normalized genre, $\sigma$ is source, $t$ is document text, $L(t)$ is token length, $k(t)$ is the length bucket, and $\mathbf{m}$ contains optional metadata. Author identifiers are hashed deterministically as}
\[
{\color{editred}a=\mathrm{SHA256}(\sigma : \texttt{raw\_author}).}
\]
{\color{editred}Token length is computed with a fixed tokenizer $T(\cdot)$ (here, \texttt{cl100k\_base}):}
\[
{\color{editred}L(t)=|T(t)|.}
\]
{\color{editred}Length buckets are then assigned by}
\[
{\color{editred}
k(t)=
\begin{cases}
\texttt{short} & 1 \le L(t) \le 10,\\
\texttt{medium} & 11 \le L(t) \le 100,\\
\texttt{long} & 101 \le L(t) \le 500,\\
\texttt{extra\_long} & L(t) > 500.
\end{cases}
}
\]
{\color{editred}During ingestion, the builder may apply a bounded buffer shuffle to each dataset stream, which approximates random ordering without loading the full source into memory.}

\subsection{Stage 1: Build \& Normalization - Chunking and Author Qualification}
\label{subsec:construct-struct}
{\color{editred}The first stage prepares a large author-qualified pool before any final size cap is enforced. If a raw document exceeds the chunking threshold, the pipeline segments it using paragraph and punctuation boundaries, with token-aware fallback splitting for very long sentences. Given a document $d$ with text $t$, chunking produces}
\[
{\color{editred}C(d)=\{t_1,\dots,t_n\}, \qquad L(t_i)\le L_{\max},}
\]
{\color{editred}where the pipeline uses configurable chunking parameters $(L_{\min},L_{\text{target}},L_{\max})$. Optional truncation is then applied by keeping the longest prefix that respects a specified token cap while preferring punctuation-aware segment boundaries.}

{\color{editred}After chunking, the builder applies a first-pass dirty-text filter (described below) and stores surviving documents in a bounded per-author reservoir. Let $\mathcal{D}_a$ be the documents observed for author $a$. Stage 1 keeps at most $M$ documents per author in memory via reservoir replacement, with default $M=5$. At finalization, authors with $|\mathcal{D}_a|<m$ are normally discarded, where the default target is $m=3$ and a fallback minimum of $2$ is allowed to recover sparse authors. Thus, the builder preferentially carries forward authors satisfying}
\[
{\color{editred}m \le |\mathcal{D}_a| \le M,}
\]
{\color{editred}with fallback admission for authors having exactly two clean documents. Stage 1 is intentionally permissive with respect to overall corpus size: it preserves as many clean, author-qualified items as possible for the later Quality Filtering and Bucket Balanced Sampling stages.}

\subsection{Stage 2: Quality Filtering}
\label{subsec:construct-filter}
{\color{editred}The second stage re-reads the author-qualified pool and performs stricter post-processing. First, the pipeline collapses letter-by-letter spacing artifacts such as ``h e l l o'' into normal words when the proportion or run length of single-letter alphabetic tokens is too high. Let $w=T_{\text{ws}}(t)$ denote whitespace-delimited tokens. The implementation measures}
\[
{\color{editred}
r_{\text{single}}(t)=\frac{1}{|w|}\sum_{j=1}^{|w|}\ind{|w_j|=1 \wedge \alpha(w_j)},
}
\]
\[
{\color{editred}
m_{\text{single}}(t)=\max \{\text{length of a consecutive run of single-letter alphabetic tokens in } w\}.
}
\]
{\color{editred}If $r_{\text{single}}(t)$ or $m_{\text{single}}(t)$ exceeds a threshold, the pipeline attempts spacing collapse; if the cleaned text still exceeds the threshold, the document is dropped.}

{\color{editred}The post-filter then re-tokenizes the cleaned text and reruns rule-based dirty filtering. Let $w=T(t)$ be the token sequence. The dirty-text heuristics include:}
\begin{itemize}
  \item {\color{editred}\textbf{Unique token ratio}}
  \[
  {\color{editred}
  r_{\text{uniq}}(t)=\frac{|\mathrm{unique}(w)|}{|w|},
  }
  \]
  {\color{editred}and the document is dropped if $r_{\text{uniq}}(t)<\tau_{\text{uniq}}$.}

  \item {\color{editred}\textbf{Symbol ratio}}
  \[
  {\color{editred}
  r_{\text{sym}}(t)=\frac{\#\{\text{symbols in } t\}}{\max(|t|_{\text{chars}},1)},
  }
  \]
  {\color{editred}where public-domain sources use an additional consecutive-symbol check.}

  \item {\color{editred}\textbf{Maximum repeated-character run}} {\color{editred}$m_{\text{rep}}(t)$, the longest run of the same non-space character, which is dropped when $m_{\text{rep}}(t)>K_{\text{rep}}$.}

  \item {\color{editred}\textbf{Maximum consecutive-symbol run}} {\color{editred}$m_{\text{sym}}(t)$ for punctuation-heavy public-domain sources, dropped when $m_{\text{sym}}(t)>K_{\text{sym}}$.}
\end{itemize}

{\color{editred}The pipeline also applies an ``untranslatable'' filter that removes low-information or script-mismatched text. Define the alphabetic character ratio}
\[
{\color{editred}
r_{\alpha}(t)=\frac{\#\{\text{alphabetic characters in } t\}}{\#\{\text{non-space characters in } t\}},
}
\]
\[
{\color{editred}
r_{\text{tok}}(t)=\frac{1}{|w|}\sum_{j=1}^{|w|}\ind{\alpha(w_j)},
}
\]
{\color{editred}where $\alpha(w_j)$ indicates that token $w_j$ contains at least one alphabetic character. Let $S_{\ell}$ denote the expected Unicode script set for language $\ell$. The script-match ratio is}
\[
{\color{editred}
r_{\text{script}}(t;\ell)=
\frac{\#\{\text{letters in } t \text{ whose script}\in S_{\ell}\}}
{\#\{\text{letters in } t\}}.
}
\]
{\color{editred}The post-filter rejects texts with low $r_{\alpha}(t)$, low $r_{\text{tok}}(t)$, excessive single-letter density, or low $r_{\text{script}}(t;\ell)$ once enough alphabetic characters are present. When script evidence is weak, the filter can also consult \texttt{langdetect}; if the detected language is incompatible with the target label, the document is removed at this stage.}

\subsection{Stage 3: Redundancy Reduction}
\label{subsec:construct-dedup}
{\color{editred}After Quality Filtering, the pipeline removes redundancy in three passes.}

\paragraph{Exact normalized-text duplicates.}
{\color{editred}For each document, let $\widetilde{t}=N(t)$ denote case-folded text with canonicalized whitespace. We compute a stable exact-text key}
\[
{\color{editred}
h_{\text{exact}}(t)=\mathrm{BLAKE2b}_{128}(N(t)),
}
\]
{\color{editred}and keep only the first document for each unique key.}

\paragraph{Near-text duplicates.}
{\color{editred}For documents with at least a minimum token count, we compute a 64-bit SimHash over unigram and bigram features. Let $\mathcal{F}(t)$ be the multiset of unigram and bigram features extracted from $N(t)$. The near-text signature is}
\[
{\color{editred}
h_{\text{near}}(t)=\mathrm{SimHash}_{64}(\mathcal{F}(t)).
}
\]
{\color{editred}Candidate pairs are generated by LSH banding on the 64-bit signature. Two documents $t$ and $t'$ are considered near duplicates when}
\[
{\color{editred}
\mathrm{Ham}\!\left(h_{\text{near}}(t),h_{\text{near}}(t')\right)\le
\left\lfloor (1-\theta_{\text{near}})\cdot 64 \right\rfloor,
}
\]
{\color{editred}where $\theta_{\text{near}}$ is the configured similarity threshold. By default, near-text comparisons are restricted to the same language.}

\paragraph{Near-author duplicates.}
{\color{editred}The pipeline also supports optional author-profile redundancy checks to reduce residual cross-source aliasing. For author $a$, let $\mathcal{D}_a^{(p)}$ be the first $p$ representative documents after sorting by document strength, where $p$ is a small constant. The profile text is formed by concatenation}
\[
{\color{editred}
t_a^{\star}=\bigoplus_{d\in \mathcal{D}_a^{(p)}} N(t_d),
}
\]
{\color{editred}and the profile signature is $h_{\text{author}}(a)=\mathrm{SimHash}_{64}(t_a^{\star})$. Two authors $a$ and $a'$ are treated as near duplicates when}
\[
{\color{editred}
\mathrm{Ham}\!\left(h_{\text{author}}(a),h_{\text{author}}(a')\right)\le
\left\lfloor (1-\theta_{\text{author}})\cdot 64 \right\rfloor,
}
\]
{\color{editred}subject by default to same-language and cross-source constraints. When a conflict is found, the pipeline keeps the stronger author profile, prioritizing larger document count and then larger total token mass.}

\subsection{Stage 4: Language Audit}
\label{subsec:construct-post}
{\color{editred}After Redundancy Reduction, the pipeline runs an automated language audit. For each document, it recomputes the script-match ratio $r_{\text{script}}(t;\ell)$ and marks a document as suspicious when the ratio is too low for its declared language. In addition, the pipeline samples up to a configurable number of documents for probabilistic language identification using \texttt{langdetect}. Let $\widehat{\ell}(t)$ be the top detected language and $p(\widehat{\ell}(t)\mid t)$ its confidence. A document is flagged as a high-confidence mismatch when}
\[
{\color{editred}
\widehat{\ell}(t)\neq \ell
\quad \text{and} \quad
p(\widehat{\ell}(t)\mid t)\ge \tau_{\text{conf}}.
}
\]
{\color{editred}By default, such documents are retagged rather than dropped; optional stricter settings can discard them. The audit records suspicious cases for targeted manual review together with summary counts such as mismatch rate, low-script rate, and retagged language pairs.}

\subsection{Stage 5: Bucket Balanced Sampling}
\label{subsec:construct-buckets}
{\color{editred}Let $\mathcal{D}^{\star}$ denote the audited document pool. The final benchmark target $Q$ is applied only at this stage. The pipeline first groups documents by language and assigns a target}
\[
{\color{editred}
Q_{\ell}=\mathrm{round}(Q\,p_{\ell}),
}
\]
{\color{editred}where $p_{\ell}$ is the configured language prior after renormalization over the languages that are actually available. Within language $\ell$, genre targets are computed as}
\[
{\color{editred}
Q_{\ell,g}=\mathrm{round}(Q_{\ell}\rho_{\ell,g}),
}
\]
{\color{editred}and within each language--genre slice the length-bucket targets are}
\[
{\color{editred}
Q_{\ell,g,k}=\mathrm{round}(Q_{\ell,g}\beta_k),
}
\]
{\color{editred}where $\rho_{\ell,g}$ is the configured genre weight for language $\ell$ and $\beta_k$ is the global length-bucket prior. Sampling is performed without replacement. If a bucket $(\ell,g,k)$ is underfull, the pipeline first redistributes the deficit to leftover documents from the same language and genre, and then uses a spill pool formed from remaining language-matched documents. This preserves the language target as closely as possible while relaxing finer-grained quotas only when necessary.}

\paragraph{Split construction.}
\label{subsec:construct-splits}
{\color{editred}Stage 5 also performs the final document-level stratified split. Documents are first grouped by language and then bucketed by}
\[
{\color{editred}
b(d)=(g(d),k(d)).
}
\]
{\color{editred}For each language-specific bucket $B_{\ell,g,k}$, the pipeline applies deterministic shuffling with a fixed seed and allocates integer split counts according to the requested train/dev/test ratios. If $r_s$ is the desired ratio for split $s\in\{\texttt{train},\texttt{dev},\texttt{test}\}$, the raw allocation is $|B_{\ell,g,k}|\,r_s$, floored to integers, and any leftover documents are assigned by largest fractional remainder. This preserves genre and length composition within each language while keeping the split procedure deterministic.}

{\color{editred}Because exact and near-text redundancy reduction are applied globally before splitting, duplicate leakage across train/dev/test is reduced by construction. The current benchmark does \emph{not} enforce author-disjoint splits; instead, it prioritizes balanced document distributions for downstream retrieval and verification within each split. Within each split, authors with at least two documents contribute $1$--$2$ candidate documents and $1$--$2$ query documents, and each query is paired with all candidate documents from the same author in the benchmark ground-truth records.}

{\color{editred}Overall, the pipeline yields a diverse benchmark with explicit control over language, genre, and length, while combining stream-safe ingestion, rule-based cleaning, multi-level redundancy reduction, automated language auditing, and deterministic stratified finalization.}


\subsection{Evaluation Protocol}
\label{subsec:eval-protocol}
For retrieval, each test document is used as a query and ranked against a candidate pool drawn from the same split.
\rev{For verification, we evaluate on labeled query--candidate pairs derived from the same within-split pools, computing both EER and ROC-AUC.}
\rev{By default, every non-matching candidate in the pool is treated as a negative for a given query; the toolkit also supports deterministic negative sampling for faster ablations.}
All results are computed without task-specific fine-tuning to ensure comparability across model families.
We report both aggregate performance and mandatory stratified breakdowns by language, genre, and length.
Length buckets follow the benchmark definition: short (1--10 tokens), medium (11--100), long (101--500), and extra\_long (>500).

\paragraph{Embedding extraction for LLM-based models.}
For base and instruction-tuned LLMs that do not expose a dedicated sentence-embedding head, we derive a single document vector from the final hidden states. Given an input document $x=(x_1,\dots,x_n)$, let
\[
H^{(L)}(x)=\left[h_1^{(L)},\dots,h_n^{(L)}\right], \qquad h_i^{(L)}\in\mathbb{R}^{d},
\]
denote the last-layer token representations over the non-padding positions. Our default pooled representation is the masked mean
\[
\bar{h}(x)=\frac{1}{\sum_{i=1}^{n} m_i}\sum_{i=1}^{n} m_i\, h_i^{(L)},
\]
where $m_i\in\{0,1\}$ indicates whether token $x_i$ is a valid (non-padding) token. We then L2-normalize the pooled vector,
\[
e(x)=\frac{\bar{h}(x)}{\|\bar{h}(x)\|_2},
\]
and use $e(x)$ as the document embedding for both retrieval and verification. Similarity between a query $q$ and candidate $c$ is then computed as cosine similarity, which reduces to a dot product after normalization:
\[
s(q,c)=e(q)^\top e(c).
\]
This is the default path used to obtain embeddings from generic LLM backbones in our unified pipeline. When a model family provides its own native embedding interface or explicitly recommended pooling rule (e.g., special-token or last-token pooling), we follow that model-specific implementation instead.

\subsection{Metrics}
\label{subsec:metrics}
\rev{This metric design follows standard authorship-analysis practice while making our retrieval formulation explicit: classical authorship attribution is often closed-set classification \citep{stamatatos2009survey,neal2017surveying}, recent representation-learning work frequently evaluates open-world same-author retrieval with ranking metrics \citep{riverasoto2021luar,wang2023canrep}, and authorship verification remains the pairwise same-author decision problem typically assessed with ROC-based measures \citep{pan2022av,pan2023av}.}
Let $q$ be a query with candidate set $C_q$ and positives $P_q \subset C_q$.
Let $\pi_q$ be the ranking of candidates by similarity. We define the binary
relevance at rank $i$ as
\[
\mathrm{rel}_i =
\begin{cases}
1 & \text{if } \pi_q(i) \in P_q, \\
0 & \text{otherwise}.
\end{cases}
\]
For a cutoff $K$ \rev{(we report $K=5$ in the main paper)}, we compute:

\[
\mathrm{Recall@}K(q) = \frac{1}{|P_q|} \sum_{i=1}^{K} \mathrm{rel}_i
\]
\[
\mathrm{DCG@}K(q) = \sum_{i=1}^{K} \frac{\mathrm{rel}_i}{\log_2(i+1)}
\]
\[
\mathrm{IDCG@}K(q) = \sum_{i=1}^{\min(K,|P_q|)} \frac{1}{\log_2(i+1)},
\]
\[
\mathrm{nDCG@}K(q) = \frac{\mathrm{DCG@}K(q)}{\mathrm{IDCG@}K(q)}
\]
\rededit{
\[
\mathrm{RR}(q)=
\begin{cases}
\frac{1}{\min \{ i : \mathrm{rel}_i = 1\}} & \text{if there exists a relevant item in } C_q, \\
0 & \text{otherwise.}
\end{cases}
\]
}

We define Success@K as:
\[
\mathrm{Success@}K(q) = \ind{\sum_{i=1}^{K} \mathrm{rel}_i > 0}.
\]
\rededit{We report $S@5$, $R@5$, nDCG@5, and the macro-average mean reciprocal rank (MRR) over queries.}

For authorship verification, each query--candidate pair is assigned a similarity
score $s \in \mathbb{R}$ and classified using a threshold $\tau$.
Let $\mathcal{P}$ and $\mathcal{N}$ denote the sets of positive and negative pairs,
respectively. The false acceptance rate (FAR) and false rejection rate (FRR) at
threshold $\tau$ are defined as
\begin{align}
\mathrm{FAR}(\tau) &= \frac{1}{|\mathcal{N}|} \sum_{(q,c)\in\mathcal{N}} \ind{s(q,c) \ge \tau}, \\
\mathrm{FRR}(\tau) &= \frac{1}{|\mathcal{P}|} \sum_{(q,c)\in\mathcal{P}} \ind{s(q,c) < \tau}.
\end{align}
The Equal Error Rate (EER) is defined as the operating point where the two error
rates are equal:
\[
\mathrm{EER} = \mathrm{FAR}(\tau^\ast) = \mathrm{FRR}(\tau^\ast),
\]
\[
\text{where } \tau^\ast = \arg\min_{\tau} \left| \mathrm{FAR}(\tau) - \mathrm{FRR}(\tau) \right|.
\]

{\color{editred}
The ROC curve traces $\mathrm{TPR}(\tau)=1-\mathrm{FRR}(\tau)$ against
$\mathrm{FAR}(\tau)$ as $\tau$ varies. We summarize this curve with the area
under the ROC curve (ROC-AUC):
\[
\mathrm{ROC\mbox{-}AUC} =
\frac{1}{|\mathcal{P}||\mathcal{N}|}
\sum_{(q,c^+)\in\mathcal{P}}
\sum_{(q',c^-)\in\mathcal{N}}
\left(
\ind{s(q,c^+) > s(q',c^-)}
+ \tfrac{1}{2}\ind{s(q,c^+) = s(q',c^-)}
\right).
\]
Higher ROC-AUC is better.

\paragraph{Why these metrics?}
Authorship attribution in AuthBench is an open-world retrieval problem with
potentially multiple relevant candidates per query. Success@5 captures whether
a model can surface at least one correct same-author document in a short
analyst-facing shortlist. Recall@5 measures how much of the relevant
same-author set is recovered within the top of the ranking. nDCG@5 complements
these hit-based measures by rewarding correct documents that appear earlier and
by accounting for queries with more than one relevant candidate. \rededit{MRR}
\rededit{adds a first-hit view of ranking quality, which is useful when only the first correct same-author retrieval matters operationally.}

Authorship verification is a pairwise same-author decision task. EER is useful
when a single balanced operating point is desired because it directly
summarizes the trade-off between false accepts and false rejects. ROC-AUC
complements EER by measuring score separability independent of any one
threshold, which is important when raw similarity scales differ across models.
We therefore use EER for an interpretable operating-point summary and ROC-AUC
for threshold-independent discrimination, consistent with recent authorship
verification practice \citep{pan2022av,pan2023av,wang2023canrep}.
}

\subsection{Models Evaluated}
\label{subsec:models}

{\color{editred}
\begingroup
\scriptsize
\setlength{\tabcolsep}{6pt}
\begin{longtable}{@{}p{0.27\linewidth}cp{0.53\linewidth}@{}}
\caption{Models and baselines evaluated in AuthBench. We list every system appearing in the updated leaderboard together with its model size and Hugging Face repository or baseline implementation note.}
\label{tab:models-evaluated}\\
\toprule
\textbf{Model} & \textbf{Model Size} & \textbf{Repository / Implementation} \\
\midrule
\endfirsthead
\caption[]{Models and baselines evaluated in AuthBench (continued)}\\
\toprule
\textbf{Model} & \textbf{Model Size} & \textbf{Repository / Implementation} \\
\midrule
\endhead
\midrule
\multicolumn{3}{r}{\textit{Continued on next page}} \\
\endfoot
\bottomrule
\endlastfoot
\multicolumn{3}{l}{\textit{LLMs (instruction-tuned)}} \\
\midrule
llama3-8b-instruct & 8B & meta-llama/Meta-Llama-3-8B-Instruct \\
llama3.1-8b-instruct & 8B & meta-llama/Llama-3.1-8B-Instruct \\
qwen2.5-3b-instruct & 3.1B & Qwen/Qwen2.5-3B-Instruct \\
qwen2.5-7b-instruct & 7.6B & Qwen/Qwen2.5-7B-Instruct \\
deepseek-llm-7b-chat & 7B & deepseek-ai/deepseek-llm-7b-chat \\
qwen3-4b-instruct & 4B & Qwen/Qwen3-4B-Instruct-2507 \\
\midrule
\multicolumn{3}{l}{\textit{LLMs (base)}} \\
\midrule
llama3-8b & 8B & meta-llama/Meta-Llama-3-8B \\
llama3.1-8b & 8B & meta-llama/Llama-3.1-8B \\
deepseek-llm-7b-base & 7B & deepseek-ai/deepseek-llm-7b-base \\
qwen2.5-3b & 3.1B & Qwen/Qwen2.5-3B \\
qwen3-4b & 4B & Qwen/Qwen3-4B \\
\midrule
\multicolumn{3}{l}{\textit{Embedding models (instruction-tuned)}} \\
\midrule
e5-mistral-7b-instruct & 7.1B & intfloat/e5-mistral-7b-instruct \\
gte-qwen2-7b-instruct & 7.6B & Alibaba-NLP/gte-Qwen2-7B-instruct \\
\midrule
\multicolumn{3}{l}{\textit{Embedding models}} \\
\midrule
multilingual-e5-large & 559.9M & intfloat/multilingual-e5-large \\
multilingual-e5-base & 278M & intfloat/multilingual-e5-base \\
qwen3-embedding-8b & 7.6B & Qwen/Qwen3-Embedding-8B \\
sfr-embedding-mistral & 7.1B & Salesforce/SFR-Embedding-Mistral \\
qwen3-embedding-4b & 4B & Qwen/Qwen3-Embedding-4B \\
snowflake-arctic-embed-l-v2 & 567.8M & Snowflake/snowflake-arctic-embed-l-v2.0 \\
qwen3-embedding-0.6b & 595.8M & Qwen/Qwen3-Embedding-0.6B \\
e5-large-v2 & 335.1M & intfloat/e5-large-v2 \\
e5-base-v2 & 109M & intfloat/e5-base-v2 \\
facebook-contriever & 110M & facebook/contriever \\
gte-large-en-v1.5 & 409M & Alibaba-NLP/gte-large-en-v1.5 \\
bge-m3 & 567M & BAAI/bge-m3 \\
bge-large-en-v1.5 & 335M & BAAI/bge-large-en-v1.5 \\
e5-small-v2 & 33M & intfloat/e5-small-v2 \\
gte-large & 335.1M & thenlper/gte-large \\
mxbai-embed-large-v1 & 335.1M & mixedbread-ai/mxbai-embed-large-v1 \\
bge-base-en-v1.5 & 109.5M & BAAI/bge-base-en-v1.5 \\
gte-base & 109.5M & thenlper/gte-base \\
bge-base-zh-v1.5 & 102M & BAAI/bge-base-zh-v1.5 \\
facebook-contriever-msmarco & 110M & facebook/contriever-msmarco \\
bge-large-zh-v1.5 & 326M & BAAI/bge-large-zh-v1.5 \\
distiluse-base-multilingual-cased-v2 & 134.7M & sentence-transformers/distiluse-base-multilingual-cased-v2 \\
all-roberta-large-v1 & 355.4M & sentence-transformers/all-roberta-large-v1 \\
all-mpnet-base-v2 & 109.5M & sentence-transformers/all-mpnet-base-v2 \\
bge-small-en-v1.5 & 33.4M & BAAI/bge-small-en-v1.5 \\
all-minilm-l12-v2 & 33.4M & sentence-transformers/all-MiniLM-L12-v2 \\
paraphrase-mpnet-base-v2 & 109.5M & sentence-transformers/paraphrase-mpnet-base-v2 \\
bert-base-uncased & 110.1M & bert-base-uncased \\
all-minilm-l6-v2 & 22.7M & sentence-transformers/all-MiniLM-L6-v2 \\
paraphrase-multilingual-mpnet-base-v2 & 278M & sentence-transformers/paraphrase-multilingual-mpnet-base-v2 \\
msmarco-distilbert-base-v4 & 66.4M & sentence-transformers/msmarco-distilbert-base-v4 \\
allenai-specter & 110M & allenai/specter \\
jina-embeddings-v2-small-en & 33M & jinaai/jina-embeddings-v2-small-en \\
jina-embeddings-v2-base-en & 137.4M & jinaai/jina-embeddings-v2-base-en \\
\midrule
\multicolumn{3}{l}{\textit{Lexical / non-neural baselines}} \\
\midrule
tfidf & -- & scikit-learn character 3--5 gram TF-IDF cosine baseline \\
ngram & -- & hashed character/word n-gram stylometric baseline with train-split calibrator \\
ppm & -- & fixed-order hashed character language-model approximation of PPM-style scoring \\
\end{longtable}
\endgroup
}

\paragraph{Non-neural baseline details.}
\texttt{tfidf} represents each document with scikit-learn character $3$--$5$ gram TF-IDF features and ranks candidates by cosine similarity, following the standard vector-space term-weighting formulation of \citet{salton1988termweighting}. \texttt{ngram} is a lightweight stylometric baseline inspired by the feature-based authorship verification line of \citet{koppel2004authorship}: our implementation combines hashed character $3$--$5$ grams, hashed word $1$--$2$ grams, and a small set of surface cues (length, punctuation, digits, capitalization, whitespace, and mean token length), then fits a train-split linear pair calibrator to produce same-author scores. \texttt{ppm} follows the compression-based language-modeling view of \citet{teahan2003compression}: we approximate a fixed-order character PPM scorer with hashed character counts, derive symmetric query--candidate cross-entropy features, and fit a train-split linear calibrator. For \texttt{ngram} and \texttt{ppm}, these are scalable benchmark implementations inspired by the cited methods rather than exact historical reimplementations.


\section{Full Results Tables}
\label{sec:appendix-full-results}
\rededit{This appendix reports the full zero-shot result tables for all 47 neural models and the three non-neural baselines evaluated on AuthBench. Table~\ref{tab:overall-leaderboard-full} through Table~\ref{tab:length-eer5-full} provide the complete benchmark breakdown across overall, language, genre, and length settings, while Figures~\ref{fig:overall-success5-bar}--\ref{fig:overall-eer-bar} provide metric-wise visual summaries.}

{\color{editred}\subsection{Overall Leaderboard Full Results}
This subsection reports the complete zero-shot leaderboard across all evaluated model families on the AuthBench test split.
\begingroup
\scriptsize
\setlength{\tabcolsep}{5pt}
\begin{longtable}{@{}p{0.30\linewidth}crrrrrr@{}}
\caption{Updated overall zero-shot results on AuthBench. Authorship attribution is evaluated with Success@5 (S@5), Recall@5 (R@5), nDCG@5, and MRR (higher is better). Authorship verification is evaluated with ROC-AUC (higher is better) and EER (lower is better). All 47 neural models and the three non-neural baselines are included.}
\label{tab:overall-leaderboard-full}\\
\toprule
\textbf{Model} & \textbf{Model Size} & \textbf{S@5 $\uparrow$} & \textbf{R@5 $\uparrow$} & \textbf{nDCG@5 $\uparrow$} & \textbf{MRR $\uparrow$} & \textbf{ROC-AUC $\uparrow$} & \textbf{EER $\downarrow$} \\
\midrule
\endfirsthead
\caption[]{Updated overall zero-shot results on AuthBench (continued)}\\
\toprule
\textbf{Model} & \textbf{Model Size} & \textbf{S@5 $\uparrow$} & \textbf{R@5 $\uparrow$} & \textbf{nDCG@5 $\uparrow$} & \textbf{MRR $\uparrow$} & \textbf{ROC-AUC $\uparrow$} & \textbf{EER $\downarrow$} \\
\midrule
\endhead
\midrule
\multicolumn{8}{r}{\textit{Continued on next page}} \\
\endfoot
\bottomrule
\endlastfoot
\multicolumn{8}{l}{\textit{LLMs (instruction-tuned)}} \\
\midrule
\texttt{llama3-8b-instruct} & 8B & \textbf{0.251} & \textbf{0.247} & \textbf{0.206} & \textbf{0.209} & \underline{0.967} & \underline{0.080} \\
\texttt{llama3.1-8b-instruct} & 8B & \underline{0.247} & \underline{0.243} & \underline{0.204} & \underline{0.208} & \textbf{0.968} & \textbf{0.076} \\
\texttt{qwen2.5-3b-instruct} & 3.1B & 0.213 & 0.209 & 0.172 & 0.176 & 0.962 & 0.091 \\
\texttt{qwen2.5-7b-instruct} & 7.6B & 0.207 & 0.203 & 0.168 & 0.172 & 0.964 & 0.083 \\
\texttt{deepseek-llm-7b-chat} & 7B & 0.207 & 0.204 & 0.170 & 0.174 & 0.940 & 0.128 \\
\texttt{qwen3-4b-instruct} & 4B & 0.197 & 0.193 & 0.159 & 0.163 & 0.958 & 0.094 \\
\midrule
\multicolumn{8}{l}{\textit{LLMs (base)}} \\
\midrule
\texttt{llama3-8b} & 8B & \textbf{0.254} & \textbf{0.250} & \textbf{0.210} & \textbf{0.213} & \textbf{0.967} & \textbf{0.079} \\
\texttt{llama3.1-8b} & 8B & \underline{0.253} & \underline{0.249} & \underline{0.209} & \underline{0.212} & \underline{0.967} & \underline{0.080} \\
\texttt{deepseek-llm-7b-base} & 7B & 0.220 & 0.216 & 0.180 & 0.184 & 0.954 & 0.105 \\
\texttt{qwen2.5-3b} & 3.1B & 0.212 & 0.208 & 0.172 & 0.176 & 0.962 & 0.090 \\
\texttt{qwen3-4b} & 4B & 0.210 & 0.206 & 0.171 & 0.175 & 0.960 & 0.087 \\
\midrule
\multicolumn{8}{l}{\textit{Embedding models (instruction-tuned)}} \\
\midrule
\texttt{e5-mistral-7b-instruct} & 7.1B & \textbf{0.242} & \textbf{0.238} & \textbf{0.202} & \textbf{0.205} & \underline{0.955} & \underline{0.096} \\
\texttt{gte-qwen2-7b-instruct} & 7.6B & \underline{0.240} & \underline{0.235} & \underline{0.198} & \underline{0.202} & \textbf{0.961} & \textbf{0.078} \\
\midrule
\multicolumn{8}{l}{\textit{Embedding models}} \\
\midrule
\texttt{multilingual-e5-large} & 559.9M & \textbf{0.258} & \textbf{0.254} & \textbf{0.217} & \textbf{0.220} & 0.920 & 0.157 \\
\texttt{multilingual-e5-base} & 278M & \underline{0.250} & \underline{0.245} & \underline{0.209} & \underline{0.212} & 0.918 & 0.161 \\
\texttt{qwen3-embedding-8b} & 7.6B & 0.241 & 0.237 & 0.201 & 0.203 & 0.940 & 0.135 \\
\texttt{sfr-embedding-mistral} & 7.1B & 0.240 & 0.236 & 0.201 & 0.205 & \textbf{0.956} & \textbf{0.096} \\
\texttt{qwen3-embedding-4b} & 4B & 0.236 & 0.231 & 0.196 & 0.198 & 0.953 & 0.111 \\
\texttt{snowflake-arctic-embed-l-v2} & 567.8M & 0.212 & 0.208 & 0.182 & 0.185 & 0.864 & 0.219 \\
\texttt{qwen3-embedding-0.6b} & 595.8M & 0.211 & 0.208 & 0.177 & 0.181 & \underline{0.953} & 0.108 \\
\texttt{e5-large-v2} & 335.1M & 0.208 & 0.205 & 0.173 & 0.176 & 0.887 & 0.187 \\
\texttt{e5-base-v2} & 109M & 0.200 & 0.197 & 0.165 & 0.169 & 0.888 & 0.179 \\
\texttt{facebook-contriever} & 110M & 0.195 & 0.192 & 0.160 & 0.164 & 0.895 & 0.169 \\
\texttt{gte-large-en-v1.5} & 409M & 0.189 & 0.187 & 0.156 & 0.159 & 0.923 & 0.129 \\
\texttt{bge-m3} & 567M & 0.188 & 0.185 & 0.161 & 0.164 & 0.796 & 0.281 \\
\texttt{bge-large-en-v1.5} & 335M & 0.177 & 0.175 & 0.147 & 0.150 & 0.867 & 0.192 \\
\texttt{e5-small-v2} & 33M & 0.177 & 0.174 & 0.146 & 0.150 & 0.856 & 0.216 \\
\texttt{gte-large} & 335.1M & 0.175 & 0.172 & 0.144 & 0.148 & 0.889 & 0.163 \\
\texttt{mxbai-embed-large-v1} & 335.1M & 0.174 & 0.171 & 0.143 & 0.146 & 0.865 & 0.197 \\
\texttt{bge-base-en-v1.5} & 109.5M & 0.172 & 0.169 & 0.142 & 0.145 & 0.850 & 0.211 \\
\texttt{gte-base} & 109.5M & 0.168 & 0.165 & 0.138 & 0.141 & 0.877 & 0.178 \\
\texttt{bge-base-zh-v1.5} & 102M & 0.167 & 0.164 & 0.139 & 0.143 & 0.912 & 0.152 \\
\texttt{facebook-contriever-msmarco} & 110M & 0.167 & 0.164 & 0.138 & 0.142 & 0.870 & 0.200 \\
\texttt{bge-large-zh-v1.5} & 326M & 0.163 & 0.159 & 0.135 & 0.139 & 0.903 & 0.169 \\
\texttt{\seqsplit{distiluse-base-multilingual-cased-v2}} & 134.7M & 0.162 & 0.159 & 0.134 & 0.137 & 0.836 & 0.245 \\
\texttt{all-roberta-large-v1} & 355.4M & 0.161 & 0.158 & 0.132 & 0.136 & 0.908 & 0.144 \\
\texttt{all-mpnet-base-v2} & 109.5M & 0.160 & 0.157 & 0.130 & 0.133 & 0.889 & 0.169 \\
\texttt{bge-small-en-v1.5} & 33.4M & 0.158 & 0.155 & 0.131 & 0.135 & 0.853 & 0.214 \\
\texttt{all-minilm-l12-v2} & 33.4M & 0.152 & 0.150 & 0.125 & 0.129 & 0.888 & 0.185 \\
\texttt{paraphrase-mpnet-base-v2} & 109.5M & 0.149 & 0.146 & 0.123 & 0.127 & 0.881 & 0.181 \\
\texttt{bert-base-uncased} & 110.1M & 0.145 & 0.143 & 0.118 & 0.122 & 0.944 & \underline{0.101} \\
\texttt{all-minilm-l6-v2} & 22.7M & 0.145 & 0.143 & 0.120 & 0.124 & 0.891 & 0.176 \\
\texttt{\seqsplit{paraphrase-multilingual-mpnet-base-v2}} & 278M & 0.138 & 0.136 & 0.117 & 0.119 & 0.794 & 0.281 \\
\texttt{msmarco-distilbert-base-v4} & 66.4M & 0.133 & 0.130 & 0.110 & 0.114 & 0.838 & 0.247 \\
\texttt{allenai-specter} & 110M & 0.107 & 0.105 & 0.086 & 0.094 & 0.901 & 0.169 \\
\texttt{jina-embeddings-v2-small-en} & 33M & 0.100 & 0.098 & 0.081 & 0.086 & 0.935 & 0.121 \\
\texttt{jina-embeddings-v2-base-en} & 137.4M & 0.016 & 0.016 & 0.012 & 0.014 & 0.738 & 0.328 \\
\midrule
\multicolumn{8}{l}{\textit{Lexical / non-neural baselines}} \\
\midrule
\texttt{tfidf} & -- & \textbf{0.197} & \textbf{0.194} & \textbf{0.167} & \textbf{0.183} & \underline{0.837} & \underline{0.219} \\
\texttt{ngram} & -- & \underline{0.170} & \underline{0.167} & \underline{0.145} & \underline{0.149} & \textbf{0.874} & \textbf{0.213} \\
\texttt{ppm} & -- & 0.157 & 0.156 & 0.137 & 0.142 & 0.792 & 0.298 \\
\end{longtable}
\endgroup

\subsection{Overall Metric Bar Charts}
These figures provide metric-wise visual summaries of the full leaderboard and make the relative spread between model families easier to compare at a glance.
\begin{figure}[t]
  \centering
  \safeincludegraphics[width=0.75\linewidth]{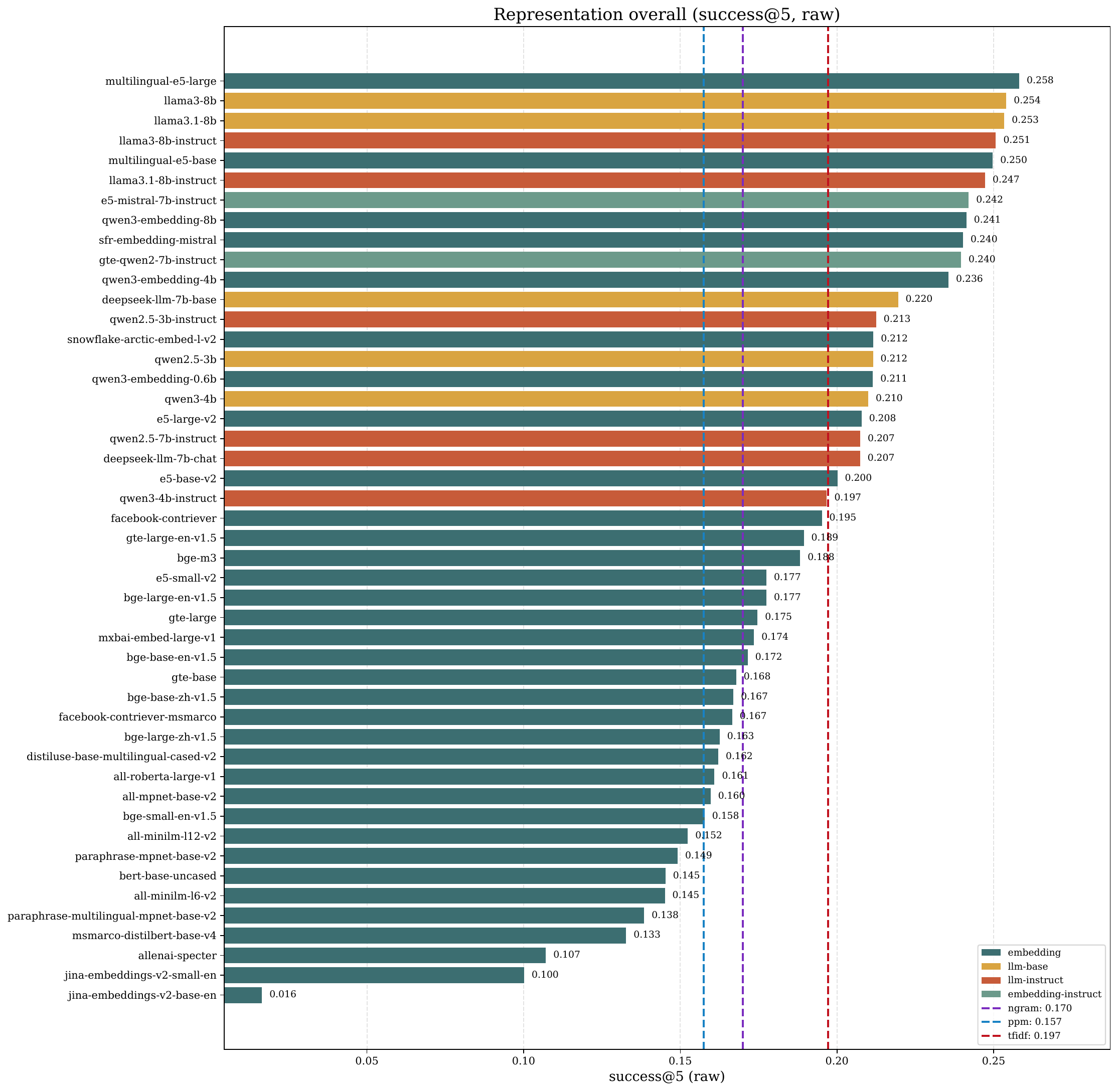}
  \caption{Overall AuthBench performance by Success@5 (S@5).}
  \label{fig:overall-success5-bar}
\end{figure}

\begin{figure}[t]
  \centering
  \safeincludegraphics[width=0.75\linewidth]{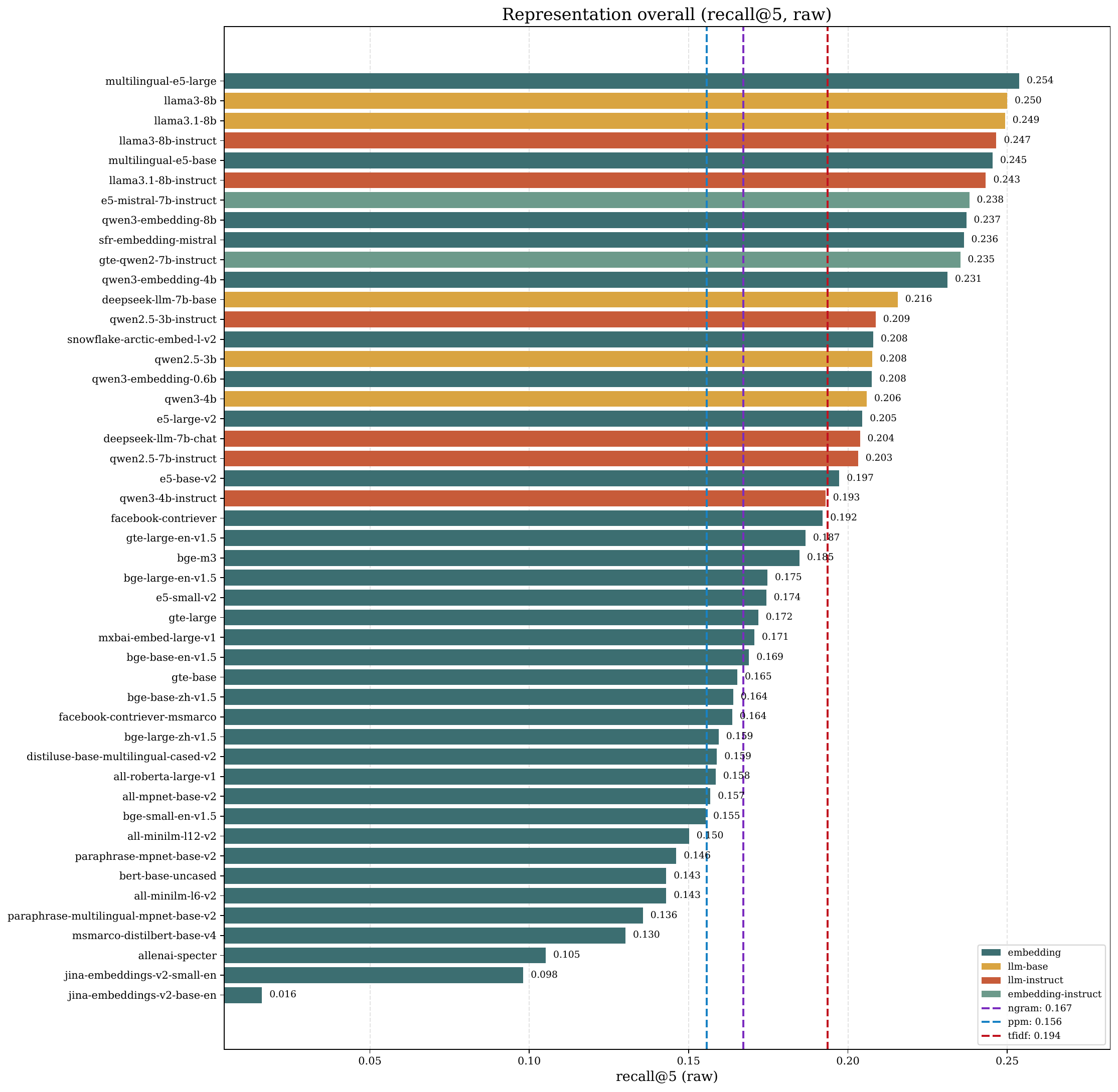}
  \caption{Overall AuthBench performance by Recall@5 (R@5).}
  \label{fig:overall-recall5-bar}
\end{figure}

\begin{figure}[t]
  \centering
  \safeincludegraphics[width=0.75\linewidth]{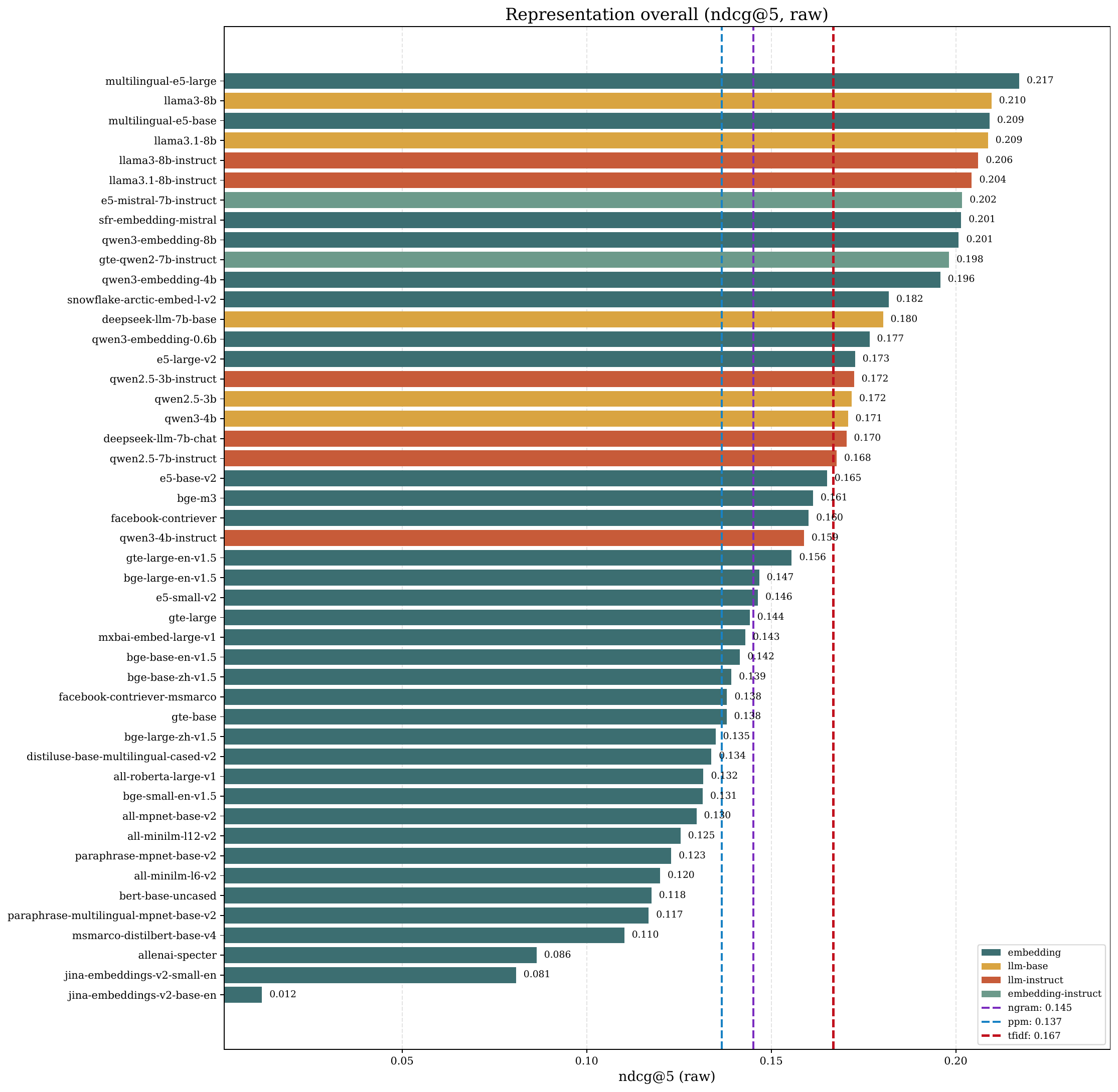}
  \caption{Overall AuthBench performance by nDCG@5.}
  \label{fig:overall-ndcg5-bar}
\end{figure}

\begin{figure}[t]
  \centering
  \safeincludegraphics[width=0.75\linewidth]{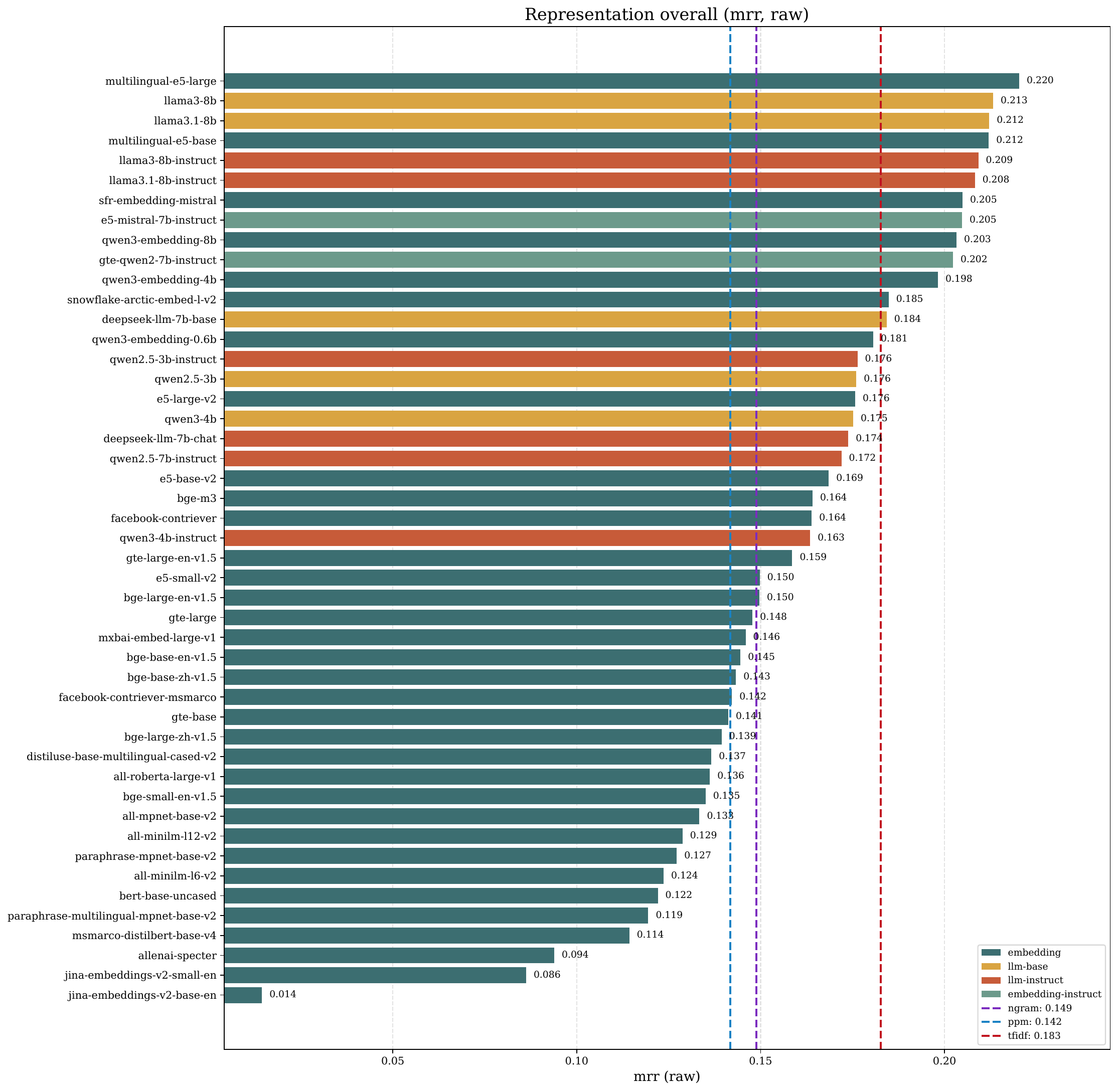}
  \caption{Overall AuthBench performance by MRR.}
  \label{fig:overall-mrr-bar}
\end{figure}

\begin{figure}[t]
  \centering
  \safeincludegraphics[width=0.75\linewidth]{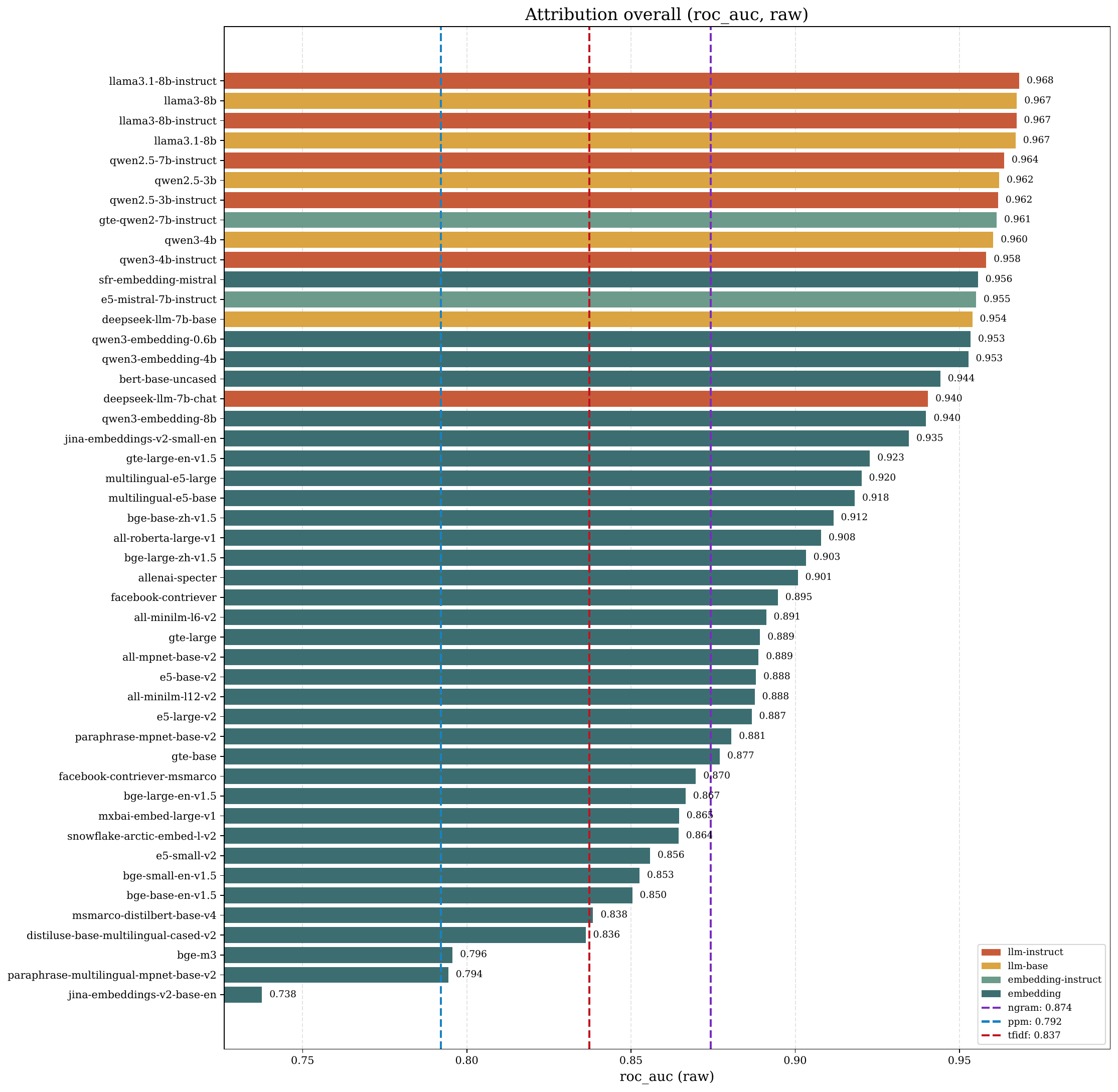}
  \caption{Overall AuthBench performance by ROC-AUC.}
  \label{fig:overall-rocauc-bar}
\end{figure}

\begin{figure}[t]
  \centering
  \safeincludegraphics[width=0.75\linewidth]{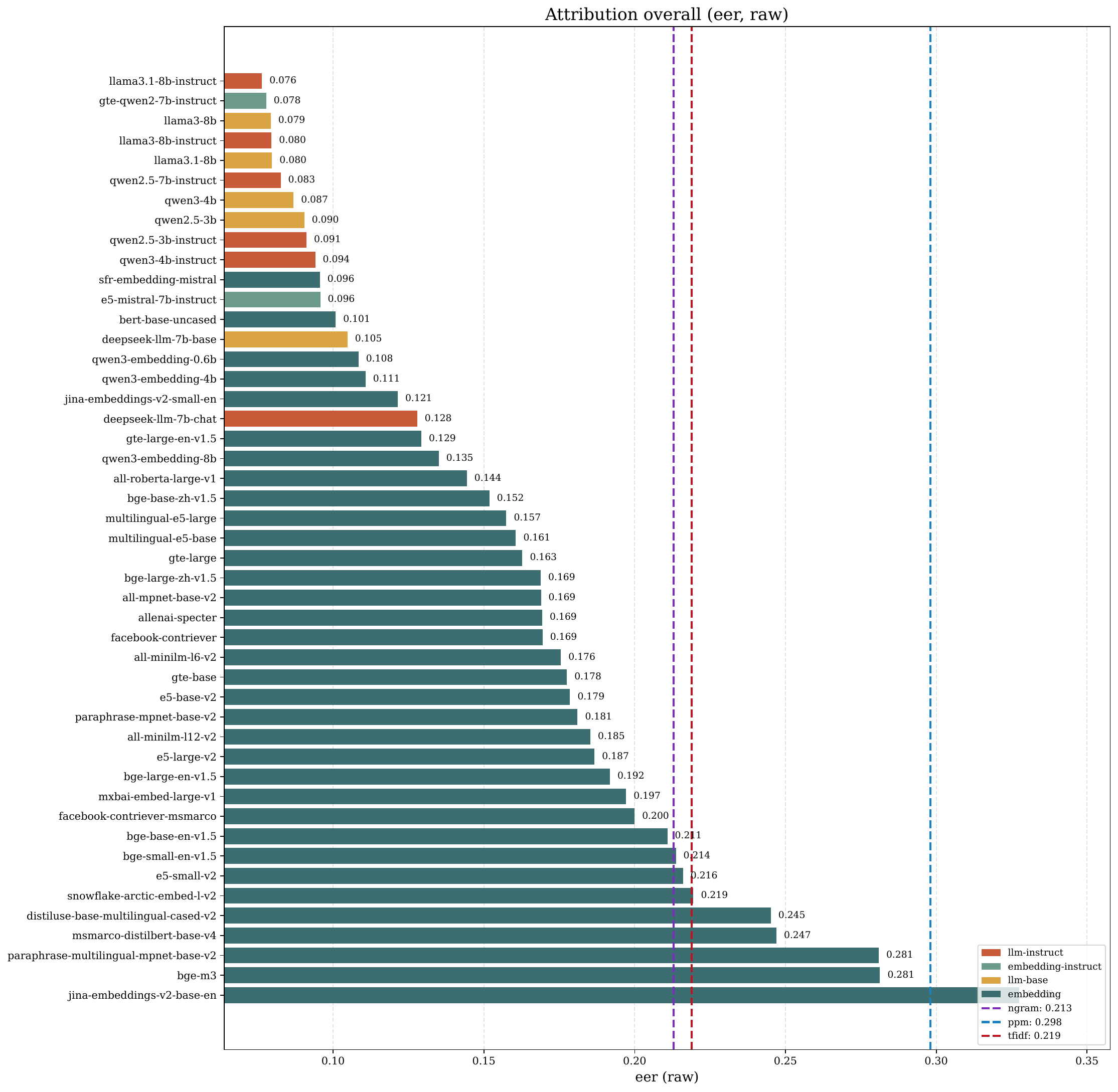}
  \caption{Overall AuthBench performance by EER.}
  \label{fig:overall-eer-bar}
\end{figure}

\subsection{Language-wise Full Results}
The following tables provide the complete language-level breakdown for both retrieval and verification-oriented evaluation.
\begingroup
\scriptsize
\setlength{\tabcolsep}{4pt}
\begin{longtable}{@{}p{0.25\linewidth}crrrrrrrrrr@{}}
\caption{Language-wise Success@5 on AuthBench (full results).}
\label{tab:lang-s5-full}\\
\toprule
\textbf{Model} & \textbf{Model Size} & \textbf{ar} & \textbf{de} & \textbf{en} & \textbf{es} & \textbf{fr} & \textbf{hi} & \textbf{ja} & \textbf{ko} & \textbf{ru} & \textbf{zh} \\
\midrule
\endfirsthead
\caption[]{Language-wise Success@5 on AuthBench (continued)}\\
\toprule
\textbf{Model} & \textbf{Model Size} & \textbf{ar} & \textbf{de} & \textbf{en} & \textbf{es} & \textbf{fr} & \textbf{hi} & \textbf{ja} & \textbf{ko} & \textbf{ru} & \textbf{zh} \\
\midrule
\endhead
\midrule
\multicolumn{12}{r}{\textit{Continued on next page}} \\
\endfoot
\bottomrule
\endlastfoot
\multicolumn{12}{l}{\textit{LLMs (instruction-tuned)}} \\
\midrule
\texttt{deepseek-llm-7b-chat} & 7B & 0.170 & 0.203 & 0.223 & 0.223 & 0.213 & 0.204 & 0.184 & 0.135 & 0.126 & 0.347 \\
\texttt{llama3-8b-instruct} & 8B & \textbf{0.226} & \underline{0.244} & \textbf{0.261} & \textbf{0.265} & \textbf{0.269} & \underline{0.289} & \textbf{0.243} & \underline{0.173} & \underline{0.164} & \textbf{0.382} \\
\texttt{llama3.1-8b-instruct} & 8B & \underline{0.211} & \textbf{0.245} & \underline{0.259} & \underline{0.265} & \underline{0.265} & \textbf{0.297} & \underline{0.229} & \textbf{0.177} & \textbf{0.165} & \underline{0.373} \\
\texttt{qwen2.5-3b-instruct} & 3.1B & 0.207 & 0.206 & 0.206 & 0.215 & 0.220 & 0.218 & 0.223 & 0.145 & 0.150 & 0.336 \\
\texttt{qwen2.5-7b-instruct} & 7.6B & 0.182 & 0.210 & 0.201 & 0.208 & 0.214 & 0.215 & 0.215 & 0.143 & 0.143 & 0.343 \\
\texttt{qwen3-4b-instruct} & 4B & 0.187 & 0.195 & 0.187 & 0.203 & 0.198 & 0.229 & 0.212 & 0.143 & 0.143 & 0.304 \\
\midrule
\multicolumn{12}{l}{\textit{LLMs (base)}} \\
\midrule
\texttt{deepseek-llm-7b-base} & 7B & 0.183 & 0.216 & 0.238 & 0.223 & 0.220 & 0.238 & 0.200 & 0.144 & 0.131 & \underline{0.367} \\
\texttt{llama3-8b} & 8B & \underline{0.221} & \underline{0.247} & \textbf{0.270} & \textbf{0.272} & \textbf{0.270} & \underline{0.289} & \underline{0.242} & \underline{0.183} & \underline{0.165} & \textbf{0.380} \\
\texttt{llama3.1-8b} & 8B & \textbf{0.225} & \textbf{0.249} & \underline{0.270} & \underline{0.272} & \underline{0.266} & \textbf{0.297} & \textbf{0.246} & \textbf{0.186} & \textbf{0.173} & 0.361 \\
\texttt{qwen2.5-3b} & 3.1B & 0.206 & 0.209 & 0.209 & 0.210 & 0.208 & 0.215 & 0.221 & 0.147 & 0.148 & 0.333 \\
\texttt{qwen3-4b} & 4B & 0.189 & 0.217 & 0.204 & 0.213 & 0.214 & 0.227 & 0.220 & 0.142 & 0.147 & 0.336 \\
\midrule
\multicolumn{12}{l}{\textit{Embedding models (instruction-tuned)}} \\
\midrule
\texttt{e5-mistral-7b-instruct} & 7.1B & \textbf{0.223} & \textbf{0.241} & \textbf{0.237} & \textbf{0.254} & \textbf{0.250} & \textbf{0.272} & \underline{0.223} & \textbf{0.164} & \underline{0.164} & \textbf{0.396} \\
\texttt{gte-qwen2-7b-instruct} & 7.6B & \underline{0.220} & \underline{0.241} & \underline{0.232} & \underline{0.235} & \underline{0.239} & \underline{0.252} & \textbf{0.249} & \underline{0.156} & \textbf{0.171} & \underline{0.396} \\
\midrule
\multicolumn{12}{l}{\textit{Embedding models}} \\
\midrule
\texttt{all-minilm-l12-v2} & 33.4M & 0.128 & 0.135 & 0.181 & 0.157 & 0.162 & 0.195 & 0.135 & 0.083 & 0.084 & 0.246 \\
\texttt{all-minilm-l6-v2} & 22.7M & 0.117 & 0.128 & 0.173 & 0.157 & 0.152 & 0.184 & 0.116 & 0.076 & 0.077 & 0.245 \\
\texttt{all-mpnet-base-v2} & 109.5M & 0.152 & 0.154 & 0.190 & 0.166 & 0.166 & 0.176 & 0.148 & 0.092 & 0.092 & 0.230 \\
\texttt{all-roberta-large-v1} & 355.4M & 0.159 & 0.180 & 0.195 & 0.195 & 0.181 & 0.198 & 0.155 & 0.091 & 0.094 & 0.170 \\
\texttt{allenai-specter} & 110M & 0.092 & 0.139 & 0.133 & 0.126 & 0.133 & 0.133 & 0.055 & 0.080 & 0.066 & 0.101 \\
\texttt{bert-base-uncased} & 110.1M & 0.135 & 0.136 & 0.188 & 0.140 & 0.155 & 0.156 & 0.112 & 0.095 & 0.079 & 0.199 \\
\texttt{bge-base-en-v1.5} & 109.5M & 0.155 & 0.171 & 0.193 & 0.200 & 0.203 & 0.195 & 0.148 & 0.099 & 0.101 & 0.241 \\
\texttt{bge-base-zh-v1.5} & 102M & 0.116 & 0.124 & 0.147 & 0.153 & 0.137 & 0.193 & 0.190 & 0.124 & 0.086 & 0.407 \\
\texttt{bge-large-en-v1.5} & 335M & 0.153 & 0.176 & 0.198 & 0.205 & 0.207 & 0.184 & 0.157 & 0.105 & 0.114 & 0.249 \\
\texttt{bge-large-zh-v1.5} & 326M & 0.113 & 0.127 & 0.144 & 0.147 & 0.131 & 0.176 & 0.175 & 0.116 & 0.085 & 0.401 \\
\texttt{bge-m3} & 567M & 0.213 & 0.159 & 0.155 & 0.174 & 0.188 & 0.176 & 0.165 & 0.138 & 0.129 & 0.369 \\
\texttt{bge-small-en-v1.5} & 33.4M & 0.131 & 0.145 & 0.184 & 0.180 & 0.175 & 0.139 & 0.132 & 0.089 & 0.087 & 0.253 \\
\texttt{\seqsplit{distiluse-base-multilingual-cased-v2}} & 134.7M & 0.178 & 0.148 & 0.160 & 0.148 & 0.142 & 0.193 & 0.139 & 0.109 & 0.111 & 0.281 \\
\texttt{e5-base-v2} & 109M & 0.182 & 0.201 & 0.227 & 0.242 & 0.236 & 0.215 & 0.194 & 0.110 & 0.128 & 0.255 \\
\texttt{e5-large-v2} & 335.1M & 0.203 & 0.197 & 0.229 & 0.248 & 0.249 & 0.210 & 0.192 & 0.120 & 0.141 & 0.269 \\
\texttt{e5-small-v2} & 33M & 0.150 & 0.178 & 0.201 & 0.199 & 0.193 & 0.207 & 0.171 & 0.092 & 0.116 & 0.254 \\
\texttt{facebook-contriever} & 110M & 0.172 & 0.187 & \textbf{0.261} & 0.228 & 0.236 & 0.246 & 0.158 & 0.116 & 0.117 & 0.200 \\
\texttt{facebook-contriever-msmarco} & 110M & 0.152 & 0.151 & 0.208 & 0.197 & 0.178 & 0.201 & 0.130 & 0.099 & 0.094 & 0.224 \\
\texttt{gte-base} & 109.5M & 0.119 & 0.176 & 0.210 & 0.206 & 0.205 & 0.125 & 0.149 & 0.080 & 0.099 & 0.224 \\
\texttt{gte-large} & 335.1M & 0.113 & 0.182 & 0.216 & 0.219 & 0.217 & 0.139 & 0.149 & 0.084 & 0.113 & 0.232 \\
\texttt{gte-large-en-v1.5} & 409M & 0.181 & 0.185 & 0.223 & 0.228 & 0.222 & 0.210 & 0.180 & 0.111 & 0.115 & 0.231 \\
\texttt{jina-embeddings-v2-base-en} & 137.4M & 0.050 & 0.008 & 0.005 & 0.005 & 0.006 & 0.065 & 0.029 & 0.013 & 0.013 & 0.027 \\
\texttt{jina-embeddings-v2-small-en} & 33M & 0.082 & 0.077 & 0.108 & 0.101 & 0.081 & 0.119 & 0.097 & 0.081 & 0.057 & 0.188 \\
\texttt{msmarco-distilbert-base-v4} & 66.4M & 0.123 & 0.113 & 0.157 & 0.127 & 0.115 & 0.142 & 0.102 & 0.091 & 0.079 & 0.224 \\
\texttt{multilingual-e5-base} & 278M & \underline{0.251} & 0.210 & 0.226 & \underline{0.267} & \underline{0.262} & 0.238 & \underline{0.253} & \underline{0.185} & \underline{0.188} & 0.413 \\
\texttt{multilingual-e5-large} & 559.9M & \textbf{0.252} & 0.221 & 0.231 & \textbf{0.284} & \textbf{0.280} & \textbf{0.263} & \textbf{0.268} & \textbf{0.196} & \textbf{0.190} & \underline{0.422} \\
\texttt{mxbai-embed-large-v1} & 335.1M & 0.143 & 0.173 & 0.196 & 0.198 & 0.204 & 0.193 & 0.154 & 0.101 & 0.116 & 0.240 \\
\texttt{paraphrase-mpnet-base-v2} & 109.5M & 0.134 & 0.136 & 0.182 & 0.163 & 0.138 & 0.210 & 0.139 & 0.098 & 0.091 & 0.203 \\
\texttt{\seqsplit{paraphrase-multilingual-mpnet-base-v2}} & 278M & 0.151 & 0.111 & 0.124 & 0.127 & 0.123 & 0.147 & 0.123 & 0.091 & 0.102 & 0.269 \\
\texttt{qwen3-embedding-0.6b} & 595.8M & 0.208 & 0.192 & 0.188 & 0.207 & 0.200 & 0.221 & 0.219 & 0.147 & 0.139 & 0.399 \\
\texttt{qwen3-embedding-4b} & 4B & 0.220 & 0.231 & 0.213 & 0.227 & 0.243 & 0.218 & 0.241 & 0.167 & 0.160 & 0.422 \\
\texttt{qwen3-embedding-8b} & 7.6B & 0.239 & \textbf{0.242} & 0.209 & 0.239 & 0.241 & 0.244 & 0.251 & 0.169 & 0.170 & \textbf{0.425} \\
\texttt{sfr-embedding-mistral} & 7.1B & 0.221 & \underline{0.236} & \underline{0.236} & 0.255 & 0.248 & \underline{0.258} & 0.214 & 0.162 & 0.162 & 0.397 \\
\texttt{snowflake-arctic-embed-l-v2} & 567.8M & 0.240 & 0.161 & 0.184 & 0.187 & 0.202 & 0.252 & 0.215 & 0.154 & 0.160 & 0.385 \\
\midrule
\multicolumn{12}{l}{\textit{Lexical / non-neural baselines}} \\
\midrule
\texttt{ngram} & -- & \underline{0.238} & 0.137 & 0.126 & 0.162 & \underline{0.166} & 0.244 & \underline{0.160} & \underline{0.168} & \underline{0.132} & \underline{0.267} \\
\texttt{ppm} & -- & 0.229 & \underline{0.140} & \textbf{0.149} & \underline{0.190} & \textbf{0.172} & \underline{0.255} & 0.113 & 0.103 & 0.118 & 0.184 \\
\texttt{tfidf} & -- & \textbf{0.254} & \textbf{0.148} & \underline{0.146} & \textbf{0.190} & 0.155 & \textbf{0.280} & \textbf{0.226} & \textbf{0.184} & \textbf{0.154} & \textbf{0.346} \\
\end{longtable}
\endgroup

\begingroup
\scriptsize
\setlength{\tabcolsep}{4pt}
\begin{longtable}{@{}p{0.25\linewidth}crrrrrrrrrr@{}}
\caption{Language-wise EER on AuthBench (full results).}
\label{tab:lang-eer5-full}\\
\toprule
\textbf{Model} & \textbf{Model Size} & \textbf{ar} & \textbf{de} & \textbf{en} & \textbf{es} & \textbf{fr} & \textbf{hi} & \textbf{ja} & \textbf{ko} & \textbf{ru} & \textbf{zh} \\
\midrule
\endfirsthead
\caption[]{Language-wise EER on AuthBench (continued)}\\
\toprule
\textbf{Model} & \textbf{Model Size} & \textbf{ar} & \textbf{de} & \textbf{en} & \textbf{es} & \textbf{fr} & \textbf{hi} & \textbf{ja} & \textbf{ko} & \textbf{ru} & \textbf{zh} \\
\midrule
\endhead
\midrule
\multicolumn{12}{r}{\textit{Continued on next page}} \\
\endfoot
\bottomrule
\endlastfoot
\multicolumn{12}{l}{\textit{LLMs (instruction-tuned)}} \\
\midrule
\texttt{deepseek-llm-7b-chat} & 7B & 0.125 & 0.103 & 0.141 & 0.089 & 0.103 & 0.071 & 0.061 & 0.081 & 0.123 & 0.074 \\
\texttt{llama3-8b-instruct} & 8B & \underline{0.073} & \underline{0.064} & \underline{0.099} & \textbf{0.064} & \textbf{0.061} & 0.051 & \textbf{0.042} & \textbf{0.054} & \textbf{0.112} & \underline{0.057} \\
\texttt{llama3.1-8b-instruct} & 8B & 0.073 & \textbf{0.062} & \textbf{0.096} & \underline{0.066} & \underline{0.062} & \textbf{0.045} & \underline{0.042} & \underline{0.055} & \underline{0.114} & 0.059 \\
\texttt{qwen2.5-3b-instruct} & 3.1B & \textbf{0.072} & 0.072 & 0.131 & 0.074 & 0.068 & 0.054 & 0.043 & 0.060 & 0.119 & 0.062 \\
\texttt{qwen2.5-7b-instruct} & 7.6B & 0.075 & 0.070 & 0.118 & 0.068 & 0.064 & \underline{0.048} & 0.044 & 0.059 & 0.120 & \textbf{0.056} \\
\texttt{qwen3-4b-instruct} & 4B & 0.080 & 0.071 & 0.131 & 0.084 & 0.078 & 0.057 & 0.046 & 0.063 & 0.121 & 0.061 \\
\midrule
\multicolumn{12}{l}{\textit{LLMs (base)}} \\
\midrule
\texttt{deepseek-llm-7b-base} & 7B & 0.082 & 0.072 & 0.128 & 0.076 & 0.071 & 0.062 & 0.043 & 0.061 & 0.121 & 0.065 \\
\texttt{llama3-8b} & 8B & \underline{0.074} & \textbf{0.062} & \textbf{0.094} & \underline{0.064} & \underline{0.061} & \underline{0.051} & \textbf{0.041} & \textbf{0.052} & \underline{0.110} & \textbf{0.056} \\
\texttt{llama3.1-8b} & 8B & \textbf{0.073} & \underline{0.062} & \underline{0.094} & \textbf{0.064} & \textbf{0.060} & \textbf{0.045} & \underline{0.042} & \underline{0.053} & \textbf{0.109} & 0.060 \\
\texttt{qwen2.5-3b} & 3.1B & 0.074 & 0.072 & 0.129 & 0.074 & 0.069 & 0.057 & 0.043 & 0.060 & 0.119 & 0.060 \\
\texttt{qwen3-4b} & 4B & 0.079 & 0.066 & 0.123 & 0.077 & 0.068 & 0.057 & 0.045 & 0.064 & 0.123 & \underline{0.059} \\
\midrule
\multicolumn{12}{l}{\textit{Embedding models (instruction-tuned)}} \\
\midrule
\texttt{e5-mistral-7b-instruct} & 7.1B & \textbf{0.072} & \underline{0.072} & \underline{0.113} & \textbf{0.067} & \underline{0.064} & \underline{0.048} & \underline{0.043} & \underline{0.063} & \textbf{0.118} & \underline{0.061} \\
\texttt{gte-qwen2-7b-instruct} & 7.6B & \underline{0.073} & \textbf{0.069} & \textbf{0.107} & \underline{0.067} & \textbf{0.063} & \textbf{0.045} & \textbf{0.041} & \textbf{0.061} & \underline{0.118} & \textbf{0.054} \\
\midrule
\multicolumn{12}{l}{\textit{Embedding models}} \\
\midrule
\texttt{all-minilm-l12-v2} & 33.4M & 0.085 & 0.207 & 0.219 & 0.107 & 0.123 & 0.062 & 0.120 & 0.105 & 0.122 & 0.102 \\
\texttt{all-minilm-l6-v2} & 22.7M & 0.088 & 0.156 & 0.217 & 0.099 & 0.109 & 0.054 & 0.097 & 0.099 & \textbf{0.114} & 0.103 \\
\texttt{all-mpnet-base-v2} & 109.5M & 0.081 & 0.132 & 0.226 & 0.096 & 0.101 & 0.051 & 0.095 & 0.087 & 0.115 & 0.107 \\
\texttt{all-roberta-large-v1} & 355.4M & 0.082 & 0.083 & 0.192 & \underline{0.072} & 0.082 & 0.042 & 0.082 & 0.065 & 0.117 & 0.096 \\
\texttt{allenai-specter} & 110M & 0.188 & 0.147 & 0.208 & 0.167 & 0.168 & 0.210 & 0.181 & 0.170 & 0.132 & 0.164 \\
\texttt{bert-base-uncased} & 110.1M & 0.084 & \underline{0.074} & 0.151 & 0.078 & 0.076 & 0.054 & 0.070 & 0.070 & 0.124 & 0.106 \\
\texttt{bge-base-en-v1.5} & 109.5M & 0.082 & 0.140 & 0.309 & 0.144 & 0.135 & 0.049 & 0.090 & 0.074 & 0.129 & 0.129 \\
\texttt{bge-base-zh-v1.5} & 102M & 0.103 & 0.131 & 0.190 & 0.116 & 0.152 & 0.096 & 0.063 & 0.070 & 0.124 & 0.172 \\
\texttt{bge-large-en-v1.5} & 335M & 0.079 & 0.115 & 0.287 & 0.122 & 0.133 & 0.049 & 0.078 & 0.077 & 0.132 & 0.119 \\
\texttt{bge-large-zh-v1.5} & 326M & 0.094 & 0.173 & 0.241 & 0.164 & 0.192 & 0.126 & 0.104 & 0.099 & 0.124 & 0.205 \\
\texttt{bge-m3} & 567M & 0.265 & 0.273 & 0.286 & 0.302 & 0.265 & 0.320 & 0.275 & 0.280 & 0.330 & 0.216 \\
\texttt{bge-small-en-v1.5} & 33.4M & 0.099 & 0.130 & 0.250 & 0.159 & 0.128 & 0.068 & 0.094 & 0.116 & 0.144 & 0.140 \\
\texttt{\seqsplit{distiluse-base-multilingual-cased-v2}} & 134.7M & 0.253 & 0.256 & 0.206 & 0.262 & 0.241 & 0.271 & 0.235 & 0.238 & 0.308 & 0.210 \\
\texttt{e5-base-v2} & 109M & 0.077 & 0.118 & 0.242 & 0.116 & 0.122 & 0.071 & 0.097 & 0.082 & 0.132 & 0.107 \\
\texttt{e5-large-v2} & 335.1M & 0.099 & 0.159 & 0.244 & 0.148 & 0.136 & 0.068 & 0.139 & 0.092 & 0.169 & 0.139 \\
\texttt{e5-small-v2} & 33M & 0.100 & 0.189 & 0.278 & 0.185 & 0.177 & 0.092 & 0.149 & 0.086 & 0.167 & 0.130 \\
\texttt{facebook-contriever} & 110M & 0.076 & 0.096 & 0.177 & 0.095 & 0.095 & 0.051 & 0.052 & 0.067 & 0.118 & 0.084 \\
\texttt{facebook-contriever-msmarco} & 110M & 0.086 & 0.143 & 0.228 & 0.127 & 0.170 & 0.068 & 0.067 & 0.072 & 0.125 & 0.106 \\
\texttt{gte-base} & 109.5M & 0.079 & 0.093 & 0.264 & 0.097 & 0.092 & 0.042 & 0.068 & 0.074 & 0.127 & 0.113 \\
\texttt{gte-large} & 335.1M & 0.079 & 0.090 & 0.263 & 0.091 & 0.086 & \textbf{0.037} & 0.061 & 0.075 & 0.124 & 0.113 \\
\texttt{gte-large-en-v1.5} & 409M & 0.081 & 0.081 & 0.216 & 0.073 & \underline{0.068} & \underline{0.040} & 0.078 & 0.109 & 0.126 & 0.123 \\
\texttt{jina-embeddings-v2-base-en} & 137.4M & 0.201 & 0.345 & 0.376 & 0.349 & 0.352 & 0.178 & 0.289 & 0.350 & 0.287 & 0.252 \\
\texttt{jina-embeddings-v2-small-en} & 33M & 0.081 & 0.142 & 0.164 & 0.136 & 0.144 & 0.057 & 0.078 & 0.071 & 0.119 & 0.094 \\
\texttt{msmarco-distilbert-base-v4} & 66.4M & 0.083 & 0.118 & 0.249 & 0.111 & 0.122 & 0.076 & 0.115 & 0.083 & 0.131 & 0.119 \\
\texttt{multilingual-e5-base} & 278M & 0.140 & 0.143 & 0.201 & 0.170 & 0.137 & 0.213 & 0.096 & 0.098 & 0.185 & 0.078 \\
\texttt{multilingual-e5-large} & 559.9M & 0.133 & 0.136 & 0.202 & 0.147 & 0.124 & 0.190 & 0.089 & 0.106 & 0.176 & \underline{0.073} \\
\texttt{mxbai-embed-large-v1} & 335.1M & 0.080 & 0.124 & 0.285 & 0.130 & 0.139 & 0.051 & 0.073 & 0.074 & 0.127 & 0.124 \\
\texttt{paraphrase-mpnet-base-v2} & 109.5M & 0.081 & 0.103 & 0.221 & 0.087 & 0.088 & 0.099 & 0.104 & 0.088 & 0.127 & 0.121 \\
\texttt{\seqsplit{paraphrase-multilingual-mpnet-base-v2}} & 278M & 0.225 & 0.274 & 0.298 & 0.301 & 0.278 & 0.278 & 0.249 & 0.257 & 0.296 & 0.249 \\
\texttt{qwen3-embedding-0.6b} & 595.8M & 0.115 & 0.103 & \underline{0.148} & 0.111 & 0.072 & 0.127 & 0.049 & 0.065 & 0.123 & 0.087 \\
\texttt{qwen3-embedding-4b} & 4B & 0.092 & 0.112 & 0.156 & 0.115 & 0.091 & 0.089 & 0.051 & \underline{0.063} & 0.122 & 0.092 \\
\texttt{qwen3-embedding-8b} & 7.6B & \textbf{0.066} & 0.135 & 0.174 & 0.148 & 0.114 & 0.051 & \underline{0.045} & \textbf{0.059} & \underline{0.114} & 0.074 \\
\texttt{sfr-embedding-mistral} & 7.1B & \underline{0.071} & \textbf{0.073} & \textbf{0.112} & \textbf{0.067} & \textbf{0.064} & 0.048 & \textbf{0.043} & 0.063 & 0.117 & \textbf{0.061} \\
\texttt{snowflake-arctic-embed-l-v2} & 567.8M & 0.167 & 0.244 & 0.233 & 0.248 & 0.225 & 0.164 & 0.198 & 0.200 & 0.255 & 0.172 \\
\midrule
\multicolumn{12}{l}{\textit{Lexical / non-neural baselines}} \\
\midrule
\texttt{ngram} & -- & \underline{0.220} & \underline{0.191} & \underline{0.213} & \underline{0.194} & \underline{0.194} & \underline{0.110} & \textbf{0.134} & \textbf{0.257} & \underline{0.262} & \textbf{0.119} \\
\texttt{ppm} & -- & 0.288 & 0.262 & 0.257 & 0.257 & 0.248 & 0.278 & \underline{0.347} & 0.344 & 0.359 & \underline{0.323} \\
\texttt{tfidf} & -- & \textbf{0.136} & \textbf{0.165} & \textbf{0.200} & \textbf{0.179} & \textbf{0.188} & \textbf{0.082} & 0.347 & \underline{0.340} & \textbf{0.145} & 0.354 \\
\end{longtable}
\endgroup

\subsection{Primary-genre Full Results}
These tables report the full genre-level breakdown, highlighting how strongly authorship performance depends on discourse type.
\begingroup
\scriptsize
\setlength{\tabcolsep}{4pt}
\begin{longtable}{@{}p{0.22\linewidth}crrrrrrrrr@{}}
\caption{Primary-genre Success@5 on AuthBench (full results).}
\label{tab:genre-s5-full}\\
\toprule
\textbf{Model} & \textbf{Model Size} & \textbf{blog} & \textbf{ecomm} & \textbf{literature} & \textbf{media} & \textbf{news} & \textbf{poetry} & \textbf{qna} & \textbf{research} & \textbf{social} \\
\midrule
\endfirsthead
\caption[]{Primary-genre Success@5 on AuthBench (continued)}\\
\toprule
\textbf{Model} & \textbf{Model Size} & \textbf{blog} & \textbf{ecomm} & \textbf{literature} & \textbf{media} & \textbf{news} & \textbf{poetry} & \textbf{qna} & \textbf{research} & \textbf{social} \\
\midrule
\endhead
\midrule
\multicolumn{11}{r}{\textit{Continued on next page}} \\
\endfoot
\bottomrule
\endlastfoot
\multicolumn{11}{l}{\textit{LLMs (instruction-tuned)}} \\
\midrule
\texttt{deepseek-llm-7b-chat} & 7B & 0.196 & \underline{0.110} & 0.461 & \underline{0.053} & 0.243 & 0.556 & 0.541 & 0.713 & 0.166 \\
\texttt{llama3-8b-instruct} & 8B & \underline{0.213} & 0.107 & \textbf{0.521} & \textbf{0.070} & \textbf{0.302} & \textbf{0.639} & \textbf{0.650} & \underline{0.769} & \textbf{0.206} \\
\texttt{llama3.1-8b-instruct} & 8B & \textbf{0.220} & \textbf{0.112} & \underline{0.518} & 0.035 & \underline{0.293} & 0.583 & \underline{0.618} & \textbf{0.787} & \underline{0.204} \\
\texttt{qwen2.5-3b-instruct} & 3.1B & 0.171 & 0.100 & 0.483 & 0.035 & 0.252 & \underline{0.639} & 0.573 & 0.731 & 0.171 \\
\texttt{qwen2.5-7b-instruct} & 7.6B & 0.162 & 0.084 & 0.496 & 0.035 & 0.242 & 0.583 & 0.567 & 0.719 & 0.167 \\
\texttt{qwen3-4b-instruct} & 4B & 0.149 & 0.080 & 0.462 & 0.053 & 0.227 & 0.611 & 0.548 & 0.688 & 0.160 \\
\midrule
\multicolumn{11}{l}{\textit{LLMs (base)}} \\
\midrule
\texttt{deepseek-llm-7b-base} & 7B & 0.215 & \underline{0.114} & 0.486 & 0.035 & 0.257 & 0.639 & 0.548 & 0.719 & 0.176 \\
\texttt{llama3-8b} & 8B & \underline{0.229} & \textbf{0.119} & \underline{0.524} & \textbf{0.070} & \textbf{0.311} & \textbf{0.667} & \textbf{0.631} & \underline{0.800} & \textbf{0.207} \\
\texttt{llama3.1-8b} & 8B & \textbf{0.232} & 0.112 & \textbf{0.532} & \underline{0.053} & \underline{0.305} & \underline{0.667} & \underline{0.631} & \textbf{0.812} & \underline{0.207} \\
\texttt{qwen2.5-3b} & 3.1B & 0.178 & 0.096 & 0.487 & 0.018 & 0.248 & 0.639 & 0.573 & 0.744 & 0.170 \\
\texttt{qwen3-4b} & 4B & 0.166 & 0.084 & 0.485 & 0.035 & 0.241 & 0.639 & 0.561 & 0.738 & 0.171 \\
\midrule
\multicolumn{11}{l}{\textit{Embedding models (instruction-tuned)}} \\
\midrule
\texttt{e5-mistral-7b-instruct} & 7.1B & \textbf{0.202} & \textbf{0.110} & \underline{0.514} & \textbf{0.053} & \underline{0.262} & \underline{0.583} & \underline{0.580} & \underline{0.725} & \textbf{0.207} \\
\texttt{gte-qwen2-7b-instruct} & 7.6B & \underline{0.191} & \underline{0.091} & \textbf{0.538} & \underline{0.035} & \textbf{0.281} & \textbf{0.611} & \textbf{0.586} & \textbf{0.781} & \underline{0.197} \\
\midrule
\multicolumn{11}{l}{\textit{Embedding models}} \\
\midrule
\texttt{all-minilm-l12-v2} & 33.4M & 0.128 & 0.066 & 0.265 & 0.018 & 0.138 & 0.222 & 0.338 & 0.781 & 0.141 \\
\texttt{all-minilm-l6-v2} & 22.7M & 0.120 & 0.055 & 0.241 & 0.018 & 0.129 & 0.278 & 0.338 & 0.756 & 0.136 \\
\texttt{all-mpnet-base-v2} & 109.5M & 0.131 & 0.082 & 0.301 & 0.018 & 0.144 & 0.250 & 0.465 & \underline{0.838} & 0.144 \\
\texttt{all-roberta-large-v1} & 355.4M & 0.132 & 0.078 & 0.285 & 0.018 & 0.156 & 0.250 & 0.408 & 0.756 & 0.146 \\
\texttt{allenai-specter} & 110M & 0.093 & 0.059 & 0.265 & 0.018 & 0.097 & 0.222 & 0.255 & 0.644 & 0.089 \\
\texttt{bert-base-uncased} & 110.1M & 0.154 & 0.082 & 0.311 & 0.035 & 0.146 & 0.222 & 0.280 & 0.569 & 0.124 \\
\texttt{bge-base-en-v1.5} & 109.5M & 0.124 & 0.082 & 0.318 & 0.018 & 0.162 & 0.250 & 0.465 & \textbf{0.844} & 0.156 \\
\texttt{bge-base-zh-v1.5} & 102M & 0.082 & 0.068 & 0.310 & \underline{0.070} & 0.173 & 0.250 & 0.382 & 0.463 & 0.155 \\
\texttt{bge-large-en-v1.5} & 335M & 0.143 & 0.073 & 0.342 & 0.018 & 0.161 & 0.306 & 0.522 & 0.838 & 0.161 \\
\texttt{bge-large-zh-v1.5} & 326M & 0.082 & 0.073 & 0.305 & 0.035 & 0.167 & 0.250 & 0.408 & 0.431 & 0.151 \\
\texttt{bge-m3} & 567M & 0.108 & 0.066 & 0.419 & 0.018 & 0.146 & 0.389 & 0.433 & 0.594 & 0.181 \\
\texttt{bge-small-en-v1.5} & 33.4M & 0.116 & 0.075 & 0.273 & 0.018 & 0.140 & 0.194 & 0.490 & 0.800 & 0.147 \\
\texttt{\seqsplit{distiluse-base-multilingual-cased-v2}} & 134.7M & 0.104 & 0.080 & 0.326 & 0.000 & 0.149 & 0.472 & 0.497 & 0.600 & 0.148 \\
\texttt{e5-base-v2} & 109M & 0.151 & 0.098 & 0.369 & 0.018 & 0.212 & 0.333 & 0.592 & 0.800 & 0.176 \\
\texttt{e5-large-v2} & 335.1M & 0.155 & 0.089 & 0.388 & 0.018 & 0.216 & 0.306 & \underline{0.643} & 0.819 & 0.184 \\
\texttt{e5-small-v2} & 33M & 0.124 & 0.098 & 0.338 & 0.035 & 0.182 & 0.306 & 0.567 & 0.750 & 0.156 \\
\texttt{facebook-contriever} & 110M & \underline{0.193} & \textbf{0.114} & 0.380 & 0.018 & 0.237 & 0.361 & 0.433 & 0.744 & 0.159 \\
\texttt{facebook-contriever-msmarco} & 110M & 0.154 & 0.094 & 0.292 & 0.018 & 0.171 & 0.250 & 0.439 & 0.738 & 0.147 \\
\texttt{gte-base} & 109.5M & 0.143 & 0.084 & 0.320 & 0.018 & 0.167 & 0.194 & 0.522 & 0.831 & 0.146 \\
\texttt{gte-large} & 335.1M & 0.144 & 0.068 & 0.343 & 0.018 & 0.171 & 0.278 & 0.529 & 0.831 & 0.154 \\
\texttt{gte-large-en-v1.5} & 409M & 0.141 & 0.082 & 0.351 & 0.018 & 0.208 & 0.278 & 0.561 & 0.806 & 0.165 \\
\texttt{jina-embeddings-v2-base-en} & 137.4M & 0.005 & 0.002 & 0.028 & 0.000 & 0.012 & 0.000 & 0.006 & 0.006 & 0.018 \\
\texttt{jina-embeddings-v2-small-en} & 33M & 0.063 & 0.041 & 0.194 & 0.018 & 0.089 & 0.083 & 0.127 & 0.338 & 0.096 \\
\texttt{msmarco-distilbert-base-v4} & 66.4M & 0.108 & 0.057 & 0.225 & 0.035 & 0.135 & 0.167 & 0.293 & 0.625 & 0.120 \\
\texttt{multilingual-e5-base} & 278M & 0.146 & \underline{0.107} & 0.453 & 0.053 & \underline{0.258} & 0.500 & 0.631 & 0.756 & \underline{0.230} \\
\texttt{multilingual-e5-large} & 559.9M & 0.160 & 0.100 & 0.479 & 0.053 & \textbf{0.268} & 0.444 & \textbf{0.682} & 0.769 & \textbf{0.236} \\
\texttt{mxbai-embed-large-v1} & 335.1M & 0.134 & 0.064 & 0.345 & 0.018 & 0.158 & 0.167 & 0.522 & 0.831 & 0.157 \\
\texttt{paraphrase-mpnet-base-v2} & 109.5M & 0.146 & 0.094 & 0.274 & 0.018 & 0.140 & 0.278 & 0.414 & 0.694 & 0.132 \\
\texttt{\seqsplit{paraphrase-multilingual-mpnet-base-v2}} & 278M & 0.081 & 0.053 & 0.295 & 0.000 & 0.101 & 0.389 & 0.433 & 0.644 & 0.132 \\
\texttt{qwen3-embedding-0.6b} & 595.8M & 0.124 & 0.078 & 0.461 & 0.053 & 0.211 & 0.556 & 0.529 & 0.787 & 0.188 \\
\texttt{qwen3-embedding-4b} & 4B & 0.151 & 0.084 & \underline{0.535} & 0.070 & 0.243 & \textbf{0.583} & 0.561 & 0.831 & 0.205 \\
\texttt{qwen3-embedding-8b} & 7.6B & 0.151 & 0.091 & \textbf{0.536} & \textbf{0.088} & 0.256 & 0.556 & 0.586 & 0.838 & 0.209 \\
\texttt{sfr-embedding-mistral} & 7.1B & \textbf{0.199} & 0.105 & 0.512 & 0.053 & 0.257 & \underline{0.583} & 0.567 & 0.756 & 0.207 \\
\texttt{snowflake-arctic-embed-l-v2} & 567.8M & 0.129 & 0.057 & 0.418 & 0.035 & 0.178 & 0.472 & 0.541 & 0.738 & 0.202 \\
\midrule
\multicolumn{11}{l}{\textit{Lexical / non-neural baselines}} \\
\midrule
\texttt{ngram} & -- & 0.066 & 0.050 & \underline{0.289} & \underline{0.018} & \underline{0.161} & \underline{0.278} & 0.325 & 0.431 & \underline{0.168} \\
\texttt{ppm} & -- & \textbf{0.090} & \textbf{0.064} & 0.230 & 0.000 & 0.149 & 0.250 & \underline{0.420} & \textbf{0.631} & 0.153 \\
\texttt{tfidf} & -- & \underline{0.089} & \underline{0.053} & \textbf{0.323} & \textbf{0.035} & \textbf{0.169} & \textbf{0.444} & \textbf{0.459} & \underline{0.613} & \textbf{0.198} \\
\end{longtable}
\endgroup

\begingroup
\scriptsize
\setlength{\tabcolsep}{4pt}
\begin{longtable}{@{}p{0.22\linewidth}crrrrrrrrr@{}}
\caption{Primary-genre EER on AuthBench (full results).}
\label{tab:genre-eer5-full}\\
\toprule
\textbf{Model} & \textbf{Model Size} & \textbf{blog} & \textbf{ecomm} & \textbf{literature} & \textbf{media} & \textbf{news} & \textbf{poetry} & \textbf{qna} & \textbf{research} & \textbf{social} \\
\midrule
\endfirsthead
\caption[]{Primary-genre EER on AuthBench (continued)}\\
\toprule
\textbf{Model} & \textbf{Model Size} & \textbf{blog} & \textbf{ecomm} & \textbf{literature} & \textbf{media} & \textbf{news} & \textbf{poetry} & \textbf{qna} & \textbf{research} & \textbf{social} \\
\midrule
\endhead
\midrule
\multicolumn{11}{r}{\textit{Continued on next page}} \\
\endfoot
\bottomrule
\endlastfoot
\multicolumn{11}{l}{\textit{LLMs (instruction-tuned)}} \\
\midrule
\texttt{deepseek-llm-7b-chat} & 7B & 0.133 & 0.094 & 0.076 & 0.078 & 0.111 & 0.090 & 0.110 & 0.029 & 0.132 \\
\texttt{llama3-8b-instruct} & 8B & \textbf{0.106} & \underline{0.074} & \textbf{0.036} & 0.078 & 0.070 & 0.064 & \textbf{0.063} & \textbf{0.011} & \underline{0.081} \\
\texttt{llama3.1-8b-instruct} & 8B & \underline{0.108} & \textbf{0.064} & \underline{0.037} & 0.080 & \textbf{0.065} & 0.068 & \underline{0.067} & \underline{0.012} & \textbf{0.081} \\
\texttt{qwen2.5-3b-instruct} & 3.1B & 0.137 & 0.086 & 0.048 & 0.079 & 0.075 & \textbf{0.059} & 0.098 & 0.031 & 0.091 \\
\texttt{qwen2.5-7b-instruct} & 7.6B & 0.131 & 0.080 & 0.051 & \underline{0.075} & \underline{0.070} & \underline{0.061} & 0.087 & 0.019 & 0.085 \\
\texttt{qwen3-4b-instruct} & 4B & 0.136 & 0.096 & 0.050 & \textbf{0.073} & 0.081 & 0.080 & 0.098 & 0.046 & 0.092 \\
\midrule
\multicolumn{11}{l}{\textit{LLMs (base)}} \\
\midrule
\texttt{deepseek-llm-7b-base} & 7B & 0.135 & 0.091 & 0.070 & 0.078 & 0.103 & 0.079 & 0.099 & 0.031 & 0.105 \\
\texttt{llama3-8b} & 8B & \textbf{0.105} & \textbf{0.062} & \underline{0.036} & \textbf{0.073} & \textbf{0.070} & \underline{0.068} & \underline{0.069} & \textbf{0.009} & \textbf{0.081} \\
\texttt{llama3.1-8b} & 8B & \underline{0.108} & \underline{0.063} & \textbf{0.035} & 0.081 & \underline{0.071} & 0.072 & \textbf{0.067} & \underline{0.009} & \underline{0.082} \\
\texttt{qwen2.5-3b} & 3.1B & 0.134 & 0.085 & 0.049 & 0.077 & 0.074 & \textbf{0.065} & 0.101 & 0.031 & 0.090 \\
\texttt{qwen3-4b} & 4B & 0.128 & 0.084 & 0.046 & \underline{0.074} & 0.074 & 0.078 & 0.091 & 0.025 & 0.088 \\
\midrule
\multicolumn{11}{l}{\textit{Embedding models (instruction-tuned)}} \\
\midrule
\texttt{e5-mistral-7b-instruct} & 7.1B & \underline{0.120} & \underline{0.073} & \underline{0.055} & \underline{0.085} & \underline{0.088} & \underline{0.078} & \underline{0.090} & \underline{0.025} & \underline{0.093} \\
\texttt{gte-qwen2-7b-instruct} & 7.6B & \textbf{0.113} & \textbf{0.070} & \textbf{0.047} & \textbf{0.069} & \textbf{0.067} & \textbf{0.076} & \textbf{0.075} & \textbf{0.013} & \textbf{0.082} \\
\midrule
\multicolumn{11}{l}{\textit{Embedding models}} \\
\midrule
\texttt{all-minilm-l12-v2} & 33.4M & 0.214 & 0.267 & 0.111 & 0.138 & 0.194 & 0.063 & 0.126 & 0.025 & 0.167 \\
\texttt{all-minilm-l6-v2} & 22.7M & 0.211 & 0.263 & 0.084 & 0.133 & 0.200 & 0.081 & 0.109 & 0.025 & 0.146 \\
\texttt{all-mpnet-base-v2} & 109.5M & 0.219 & 0.277 & 0.079 & 0.138 & 0.197 & 0.059 & 0.121 & 0.019 & 0.138 \\
\texttt{all-roberta-large-v1} & 355.4M & 0.206 & 0.105 & 0.076 & 0.108 & 0.175 & \textbf{0.039} & 0.121 & 0.031 & 0.114 \\
\texttt{allenai-specter} & 110M & 0.164 & 0.172 & 0.070 & 0.192 & 0.164 & 0.078 & 0.098 & 0.031 & 0.182 \\
\texttt{bert-base-uncased} & 110.1M & \underline{0.128} & 0.126 & 0.069 & 0.137 & \underline{0.102} & 0.082 & 0.082 & 0.029 & \underline{0.101} \\
\texttt{bge-base-en-v1.5} & 109.5M & 0.230 & 0.365 & 0.078 & 0.211 & 0.298 & 0.082 & 0.136 & 0.031 & 0.175 \\
\texttt{bge-base-zh-v1.5} & 102M & 0.157 & 0.142 & 0.061 & 0.310 & 0.142 & 0.062 & 0.080 & 0.056 & 0.152 \\
\texttt{bge-large-en-v1.5} & 335M & 0.222 & 0.322 & 0.080 & 0.183 & 0.259 & 0.098 & 0.117 & \textbf{0.013} & 0.151 \\
\texttt{bge-large-zh-v1.5} & 326M & 0.211 & 0.202 & 0.066 & 0.336 & 0.154 & 0.063 & 0.086 & 0.102 & 0.162 \\
\texttt{bge-m3} & 567M & 0.257 & 0.214 & 0.076 & 0.368 & 0.265 & 0.118 & 0.109 & 0.054 & 0.304 \\
\texttt{bge-small-en-v1.5} & 33.4M & 0.202 & 0.277 & 0.099 & 0.229 & 0.264 & 0.092 & 0.148 & 0.025 & 0.178 \\
\texttt{\seqsplit{distiluse-base-multilingual-cased-v2}} & 134.7M & 0.191 & 0.145 & 0.091 & 0.391 & 0.191 & 0.122 & 0.052 & 0.046 & 0.277 \\
\texttt{e5-base-v2} & 109M & 0.215 & 0.194 & 0.072 & 0.172 & 0.217 & \underline{0.056} & 0.057 & 0.025 & 0.162 \\
\texttt{e5-large-v2} & 335.1M & 0.217 & 0.180 & 0.078 & 0.207 & 0.221 & 0.078 & 0.059 & 0.031 & 0.179 \\
\texttt{e5-small-v2} & 33M & 0.244 & 0.203 & 0.095 & 0.230 & 0.264 & 0.063 & 0.063 & 0.050 & 0.205 \\
\texttt{facebook-contriever} & 110M & 0.228 & \underline{0.094} & 0.075 & 0.111 & 0.155 & 0.081 & 0.109 & \underline{0.013} & 0.142 \\
\texttt{facebook-contriever-msmarco} & 110M & 0.234 & 0.148 & 0.090 & 0.127 & 0.211 & 0.078 & 0.095 & 0.021 & 0.179 \\
\texttt{gte-base} & 109.5M & 0.210 & 0.359 & 0.077 & 0.155 & 0.243 & 0.085 & 0.109 & 0.031 & 0.138 \\
\texttt{gte-large} & 335.1M & 0.207 & 0.358 & 0.066 & 0.153 & 0.213 & 0.084 & 0.098 & 0.025 & 0.127 \\
\texttt{gte-large-en-v1.5} & 409M & 0.193 & 0.516 & \underline{0.042} & 0.149 & 0.122 & 0.059 & 0.079 & 0.031 & 0.117 \\
\texttt{jina-embeddings-v2-base-en} & 137.4M & 0.326 & 0.352 & 0.275 & 0.230 & 0.342 & 0.255 & 0.337 & 0.294 & 0.325 \\
\texttt{jina-embeddings-v2-small-en} & 33M & 0.159 & 0.126 & 0.083 & 0.125 & 0.124 & 0.083 & 0.094 & 0.082 & 0.122 \\
\texttt{msmarco-distilbert-base-v4} & 66.4M & 0.238 & 0.205 & 0.117 & 0.146 & 0.241 & 0.089 & 0.208 & 0.025 & 0.234 \\
\texttt{multilingual-e5-base} & 278M & 0.179 & 0.132 & 0.087 & \underline{0.090} & 0.145 & 0.137 & \textbf{0.027} & 0.025 & 0.151 \\
\texttt{multilingual-e5-large} & 559.9M & 0.197 & 0.160 & 0.078 & 0.105 & 0.146 & 0.118 & \underline{0.034} & 0.022 & 0.147 \\
\texttt{mxbai-embed-large-v1} & 335.1M & 0.242 & 0.311 & 0.081 & 0.175 & 0.253 & 0.096 & 0.116 & 0.013 & 0.156 \\
\texttt{paraphrase-mpnet-base-v2} & 109.5M & 0.218 & 0.128 & 0.079 & 0.144 & 0.228 & 0.093 & 0.114 & 0.019 & 0.142 \\
\texttt{\seqsplit{paraphrase-multilingual-mpnet-base-v2}} & 278M & 0.310 & 0.212 & 0.122 & 0.368 & 0.287 & 0.176 & 0.098 & 0.025 & 0.283 \\
\texttt{qwen3-embedding-0.6b} & 595.8M & 0.137 & 0.096 & 0.044 & 0.166 & 0.103 & 0.066 & 0.063 & 0.015 & 0.109 \\
\texttt{qwen3-embedding-4b} & 4B & 0.131 & 0.095 & \textbf{0.040} & 0.174 & 0.110 & 0.061 & 0.080 & 0.013 & 0.107 \\
\texttt{qwen3-embedding-8b} & 7.6B & 0.166 & 0.097 & 0.070 & 0.120 & 0.139 & 0.070 & 0.092 & 0.025 & 0.126 \\
\texttt{sfr-embedding-mistral} & 7.1B & \textbf{0.117} & \textbf{0.074} & 0.053 & \textbf{0.088} & \textbf{0.089} & 0.088 & 0.088 & 0.025 & \textbf{0.093} \\
\texttt{snowflake-arctic-embed-l-v2} & 567.8M & 0.220 & 0.158 & 0.065 & 0.324 & 0.210 & 0.078 & 0.128 & 0.056 & 0.233 \\
\midrule
\multicolumn{11}{l}{\textit{Lexical / non-neural baselines}} \\
\midrule
\texttt{ngram} & -- & \underline{0.195} & \underline{0.234} & \textbf{0.128} & \textbf{0.134} & \textbf{0.146} & \underline{0.127} & \underline{0.133} & 0.087 & \textbf{0.231} \\
\texttt{ppm} & -- & 0.214 & 0.240 & 0.275 & \underline{0.356} & 0.276 & 0.335 & 0.207 & \textbf{0.069} & 0.315 \\
\texttt{tfidf} & -- & \textbf{0.154} & \textbf{0.185} & \underline{0.148} & 0.498 & \underline{0.199} & \textbf{0.098} & \textbf{0.079} & \underline{0.075} & \underline{0.243} \\
\end{longtable}
\endgroup

\subsection{Length-bucket Full Results}
These tables give the complete length-wise breakdown and make the sparse-text versus long-document gap explicit across all evaluated systems.
\begingroup
\scriptsize
\setlength{\tabcolsep}{5pt}
\begin{longtable}{@{}p{0.30\linewidth}crrrr@{}}
\caption{Length-bucket Success@5 on AuthBench (full results).}
\label{tab:length-s5-full}\\
\toprule
\textbf{Model} & \textbf{Model Size} & \textbf{short} & \textbf{medium} & \textbf{long} & \textbf{extra\_long} \\
\midrule
\endfirsthead
\caption[]{Length-bucket Success@5 on AuthBench (continued)}\\
\toprule
\textbf{Model} & \textbf{Model Size} & \textbf{short} & \textbf{medium} & \textbf{long} & \textbf{extra\_long} \\
\midrule
\endhead
\midrule
\multicolumn{6}{r}{\textit{Continued on next page}} \\
\endfoot
\bottomrule
\endlastfoot
\multicolumn{6}{l}{\textit{LLMs (instruction-tuned)}} \\
\midrule
\texttt{deepseek-llm-7b-chat} & 7B & 0.106 & 0.161 & 0.371 & 0.524 \\
\texttt{llama3-8b-instruct} & 8B & \textbf{0.150} & \textbf{0.199} & \textbf{0.432} & \textbf{0.539} \\
\texttt{llama3.1-8b-instruct} & 8B & \underline{0.142} & \underline{0.196} & \underline{0.427} & \underline{0.531} \\
\texttt{qwen2.5-3b-instruct} & 3.1B & 0.101 & 0.164 & 0.385 & 0.508 \\
\texttt{qwen2.5-7b-instruct} & 7.6B & 0.103 & 0.159 & 0.379 & 0.496 \\
\texttt{qwen3-4b-instruct} & 4B & 0.081 & 0.150 & 0.368 & 0.488 \\
\midrule
\multicolumn{6}{l}{\textit{LLMs (base)}} \\
\midrule
\texttt{deepseek-llm-7b-base} & 7B & 0.121 & 0.169 & 0.393 & \underline{0.528} \\
\texttt{llama3-8b} & 8B & \textbf{0.150} & \textbf{0.202} & \underline{0.436} & 0.524 \\
\texttt{llama3.1-8b} & 8B & \underline{0.135} & \underline{0.202} & \textbf{0.439} & \textbf{0.543} \\
\texttt{qwen2.5-3b} & 3.1B & 0.093 & 0.163 & 0.386 & 0.520 \\
\texttt{qwen3-4b} & 4B & 0.101 & 0.161 & 0.385 & 0.496 \\
\midrule
\multicolumn{6}{l}{\textit{Embedding models (instruction-tuned)}} \\
\midrule
\texttt{e5-mistral-7b-instruct} & 7.1B & \textbf{0.161} & \textbf{0.196} & \underline{0.398} & \textbf{0.524} \\
\texttt{gte-qwen2-7b-instruct} & 7.6B & \underline{0.137} & \underline{0.192} & \textbf{0.409} & \underline{0.500} \\
\midrule
\multicolumn{6}{l}{\textit{Embedding models}} \\
\midrule
\texttt{all-minilm-l12-v2} & 33.4M & 0.128 & 0.129 & 0.222 & 0.343 \\
\texttt{all-minilm-l6-v2} & 22.7M & 0.126 & 0.122 & 0.211 & 0.335 \\
\texttt{all-mpnet-base-v2} & 109.5M & 0.122 & 0.133 & 0.246 & 0.358 \\
\texttt{all-roberta-large-v1} & 355.4M & 0.113 & 0.136 & 0.248 & 0.280 \\
\texttt{allenai-specter} & 110M & 0.069 & 0.076 & 0.206 & 0.299 \\
\texttt{bert-base-uncased} & 110.1M & 0.095 & 0.113 & 0.250 & 0.354 \\
\texttt{bge-base-en-v1.5} & 109.5M & 0.117 & 0.145 & 0.265 & 0.323 \\
\texttt{bge-base-zh-v1.5} & 102M & 0.166 & 0.132 & 0.263 & 0.386 \\
\texttt{bge-large-en-v1.5} & 335M & 0.124 & 0.149 & 0.274 & 0.346 \\
\texttt{bge-large-zh-v1.5} & 326M & 0.164 & 0.130 & 0.251 & 0.370 \\
\texttt{bge-m3} & 567M & 0.151 & 0.152 & 0.302 & 0.406 \\
\texttt{bge-small-en-v1.5} & 33.4M & 0.125 & 0.135 & 0.231 & 0.319 \\
\texttt{\seqsplit{distiluse-base-multilingual-cased-v2}} & 134.7M & 0.114 & 0.130 & 0.270 & 0.327 \\
\texttt{e5-base-v2} & 109M & 0.137 & 0.169 & 0.307 & 0.390 \\
\texttt{e5-large-v2} & 335.1M & 0.140 & 0.176 & 0.320 & 0.406 \\
\texttt{e5-small-v2} & 33M & 0.123 & 0.146 & 0.282 & 0.378 \\
\texttt{facebook-contriever} & 110M & 0.120 & 0.161 & 0.317 & 0.402 \\
\texttt{facebook-contriever-msmarco} & 110M & 0.116 & 0.140 & 0.257 & 0.366 \\
\texttt{gte-base} & 109.5M & 0.112 & 0.144 & 0.255 & 0.323 \\
\texttt{gte-large} & 335.1M & 0.118 & 0.149 & 0.262 & 0.370 \\
\texttt{gte-large-en-v1.5} & 409M & 0.108 & 0.162 & 0.293 & 0.406 \\
\texttt{jina-embeddings-v2-base-en} & 137.4M & 0.011 & 0.013 & 0.026 & 0.067 \\
\texttt{jina-embeddings-v2-small-en} & 33M & 0.088 & 0.078 & 0.161 & 0.299 \\
\texttt{msmarco-distilbert-base-v4} & 66.4M & 0.115 & 0.109 & 0.204 & 0.291 \\
\texttt{multilingual-e5-base} & 278M & \textbf{0.187} & \underline{0.213} & 0.376 & 0.417 \\
\texttt{multilingual-e5-large} & 559.9M & \underline{0.182} & \textbf{0.220} & 0.392 & 0.476 \\
\texttt{mxbai-embed-large-v1} & 335.1M & 0.116 & 0.147 & 0.267 & 0.346 \\
\texttt{paraphrase-mpnet-base-v2} & 109.5M & 0.113 & 0.119 & 0.244 & 0.335 \\
\texttt{\seqsplit{paraphrase-multilingual-mpnet-base-v2}} & 278M & 0.097 & 0.111 & 0.229 & 0.307 \\
\texttt{qwen3-embedding-0.6b} & 595.8M & 0.154 & 0.173 & 0.337 & 0.453 \\
\texttt{qwen3-embedding-4b} & 4B & 0.167 & 0.191 & 0.384 & 0.500 \\
\texttt{qwen3-embedding-8b} & 7.6B & 0.157 & 0.197 & \textbf{0.395} & \underline{0.504} \\
\texttt{sfr-embedding-mistral} & 7.1B & 0.168 & 0.193 & \underline{0.395} & \textbf{0.520} \\
\texttt{snowflake-arctic-embed-l-v2} & 567.8M & 0.161 & 0.175 & 0.329 & 0.445 \\
\midrule
\multicolumn{6}{l}{\textit{Lexical / non-neural baselines}} \\
\midrule
\texttt{ngram} & -- & \underline{0.151} & \underline{0.136} & \underline{0.271} & \underline{0.331} \\
\texttt{ppm} & -- & 0.079 & 0.134 & 0.255 & 0.240 \\
\texttt{tfidf} & -- & \textbf{0.157} & \textbf{0.164} & \textbf{0.300} & \textbf{0.437} \\
\end{longtable}
\endgroup

\begingroup
\scriptsize
\setlength{\tabcolsep}{5pt}
\begin{longtable}{@{}p{0.30\linewidth}crrrr@{}}
\caption{Length-bucket EER on AuthBench (full results).}
\label{tab:length-eer5-full}\\
\toprule
\textbf{Model} & \textbf{Model Size} & \textbf{short} & \textbf{medium} & \textbf{long} & \textbf{extra\_long} \\
\midrule
\endfirsthead
\caption[]{Length-bucket EER on AuthBench (continued)}\\
\toprule
\textbf{Model} & \textbf{Model Size} & \textbf{short} & \textbf{medium} & \textbf{long} & \textbf{extra\_long} \\
\midrule
\endhead
\midrule
\multicolumn{6}{r}{\textit{Continued on next page}} \\
\endfoot
\bottomrule
\endlastfoot
\multicolumn{6}{l}{\textit{LLMs (instruction-tuned)}} \\
\midrule
\texttt{deepseek-llm-7b-chat} & 7B & 0.185 & 0.131 & 0.096 & 0.072 \\
\texttt{llama3-8b-instruct} & 8B & \underline{0.109} & \underline{0.082} & \textbf{0.060} & 0.055 \\
\texttt{llama3.1-8b-instruct} & 8B & \textbf{0.104} & \textbf{0.078} & \underline{0.061} & \textbf{0.051} \\
\texttt{qwen2.5-3b-instruct} & 3.1B & 0.120 & 0.090 & 0.073 & 0.067 \\
\texttt{qwen2.5-7b-instruct} & 7.6B & 0.110 & 0.082 & 0.069 & \underline{0.051} \\
\texttt{qwen3-4b-instruct} & 4B & 0.153 & 0.094 & 0.075 & 0.061 \\
\midrule
\multicolumn{6}{l}{\textit{LLMs (base)}} \\
\midrule
\texttt{deepseek-llm-7b-base} & 7B & 0.161 & 0.105 & 0.084 & 0.084 \\
\texttt{llama3-8b} & 8B & \textbf{0.108} & \textbf{0.082} & \underline{0.064} & \textbf{0.054} \\
\texttt{llama3.1-8b} & 8B & \underline{0.110} & \underline{0.082} & \textbf{0.064} & \underline{0.054} \\
\texttt{qwen2.5-3b} & 3.1B & 0.119 & 0.090 & 0.072 & 0.063 \\
\texttt{qwen3-4b} & 4B & 0.128 & 0.086 & 0.070 & 0.058 \\
\midrule
\multicolumn{6}{l}{\textit{Embedding models (instruction-tuned)}} \\
\midrule
\texttt{e5-mistral-7b-instruct} & 7.1B & \underline{0.123} & \underline{0.096} & \underline{0.079} & \underline{0.083} \\
\texttt{gte-qwen2-7b-instruct} & 7.6B & \textbf{0.109} & \textbf{0.078} & \textbf{0.066} & \textbf{0.054} \\
\midrule
\multicolumn{6}{l}{\textit{Embedding models}} \\
\midrule
\texttt{all-minilm-l12-v2} & 33.4M & 0.246 & 0.197 & 0.130 & 0.123 \\
\texttt{all-minilm-l6-v2} & 22.7M & 0.248 & 0.186 & 0.121 & 0.119 \\
\texttt{all-mpnet-base-v2} & 109.5M & 0.248 & 0.176 & 0.120 & 0.144 \\
\texttt{all-roberta-large-v1} & 355.4M & 0.226 & 0.147 & 0.108 & 0.148 \\
\texttt{allenai-specter} & 110M & 0.237 & 0.178 & 0.110 & 0.094 \\
\texttt{bert-base-uncased} & 110.1M & \underline{0.147} & \underline{0.098} & \textbf{0.075} & 0.083 \\
\texttt{bge-base-en-v1.5} & 109.5M & 0.309 & 0.222 & 0.130 & 0.151 \\
\texttt{bge-base-zh-v1.5} & 102M & 0.200 & 0.143 & 0.096 & 0.089 \\
\texttt{bge-large-en-v1.5} & 335M & 0.289 & 0.204 & 0.120 & 0.148 \\
\texttt{bge-large-zh-v1.5} & 326M & 0.203 & 0.147 & 0.102 & 0.113 \\
\texttt{bge-m3} & 567M & 0.354 & 0.296 & 0.180 & 0.155 \\
\texttt{bge-small-en-v1.5} & 33.4M & 0.293 & 0.224 & 0.152 & 0.176 \\
\texttt{\seqsplit{distiluse-base-multilingual-cased-v2}} & 134.7M & 0.275 & 0.255 & 0.158 & 0.092 \\
\texttt{e5-base-v2} & 109M & 0.260 & 0.186 & 0.115 & 0.116 \\
\texttt{e5-large-v2} & 335.1M & 0.245 & 0.190 & 0.131 & 0.137 \\
\texttt{e5-small-v2} & 33M & 0.296 & 0.226 & 0.139 & 0.116 \\
\texttt{facebook-contriever} & 110M & 0.259 & 0.168 & 0.134 & 0.206 \\
\texttt{facebook-contriever-msmarco} & 110M & 0.275 & 0.202 & 0.144 & 0.184 \\
\texttt{gte-base} & 109.5M & 0.274 & 0.182 & 0.123 & 0.158 \\
\texttt{gte-large} & 335.1M & 0.251 & 0.170 & 0.109 & 0.131 \\
\texttt{gte-large-en-v1.5} & 409M & 0.200 & 0.128 & 0.095 & 0.116 \\
\texttt{jina-embeddings-v2-base-en} & 137.4M & 0.387 & 0.338 & 0.283 & 0.239 \\
\texttt{jina-embeddings-v2-small-en} & 33M & 0.175 & 0.124 & 0.087 & 0.081 \\
\texttt{msmarco-distilbert-base-v4} & 66.4M & 0.270 & 0.255 & 0.196 & 0.249 \\
\texttt{multilingual-e5-base} & 278M & 0.197 & 0.171 & 0.120 & 0.083 \\
\texttt{multilingual-e5-large} & 559.9M & 0.199 & 0.164 & 0.119 & 0.076 \\
\texttt{mxbai-embed-large-v1} & 335.1M & 0.292 & 0.209 & 0.126 & 0.143 \\
\texttt{paraphrase-mpnet-base-v2} & 109.5M & 0.260 & 0.189 & 0.124 & 0.170 \\
\texttt{\seqsplit{paraphrase-multilingual-mpnet-base-v2}} & 278M & 0.360 & 0.297 & 0.208 & 0.202 \\
\texttt{qwen3-embedding-0.6b} & 595.8M & 0.169 & 0.109 & 0.082 & \textbf{0.048} \\
\texttt{qwen3-embedding-4b} & 4B & 0.178 & 0.111 & 0.085 & \underline{0.058} \\
\texttt{qwen3-embedding-8b} & 7.6B & 0.192 & 0.134 & 0.117 & 0.094 \\
\texttt{sfr-embedding-mistral} & 7.1B & \textbf{0.121} & \textbf{0.096} & \underline{0.079} & 0.083 \\
\texttt{snowflake-arctic-embed-l-v2} & 567.8M & 0.273 & 0.235 & 0.143 & 0.072 \\
\midrule
\multicolumn{6}{l}{\textit{Lexical / non-neural baselines}} \\
\midrule
\texttt{ngram} & -- & \textbf{0.260} & \textbf{0.225} & \textbf{0.154} & \textbf{0.124} \\
\texttt{ppm} & -- & \underline{0.331} & 0.308 & 0.254 & 0.274 \\
\texttt{tfidf} & -- & 0.340 & \underline{0.225} & \underline{0.159} & \underline{0.202} \\
\end{longtable}
\endgroup
}

\section{\rededit{Post-Training Note}}
\rededit{This appendix focuses on the zero-shot benchmark results reported in the current paper. In parallel, we are conducting a broader post-training study covering multiple adaptation strategies and authorship methods, and we plan to present those findings in a second version of the paper. The zero-shot results here are intended to provide a clean foundation for that next stage of analysis.}

\section{Additional Dataset Statistics}
\label{sec:appendix-additional-stats}

\begin{figure*}[h!]
  \centering
  \safeincludegraphics[width=\textwidth]{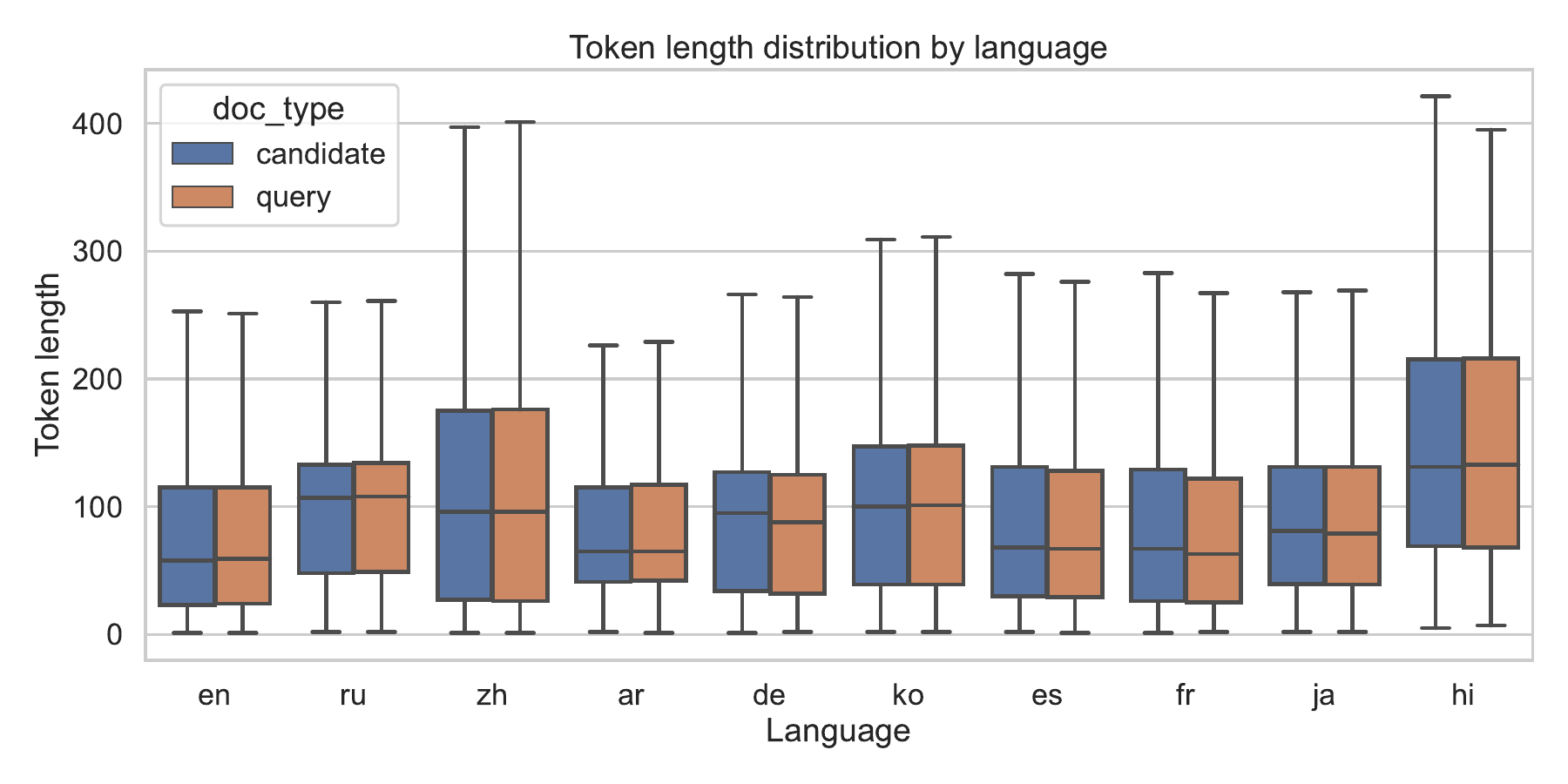}
  \caption{Token-length distribution per language. Each box plot summarizes the document-length distribution for one language in the current AuthBench release, highlighting cross-language differences in median length, spread, and long-tail behavior.}
  \label{fig:aurabench-len-by-lang}
\end{figure*}

\begin{figure}[p]
  \centering
  \safeincludegraphics[width=\textwidth]{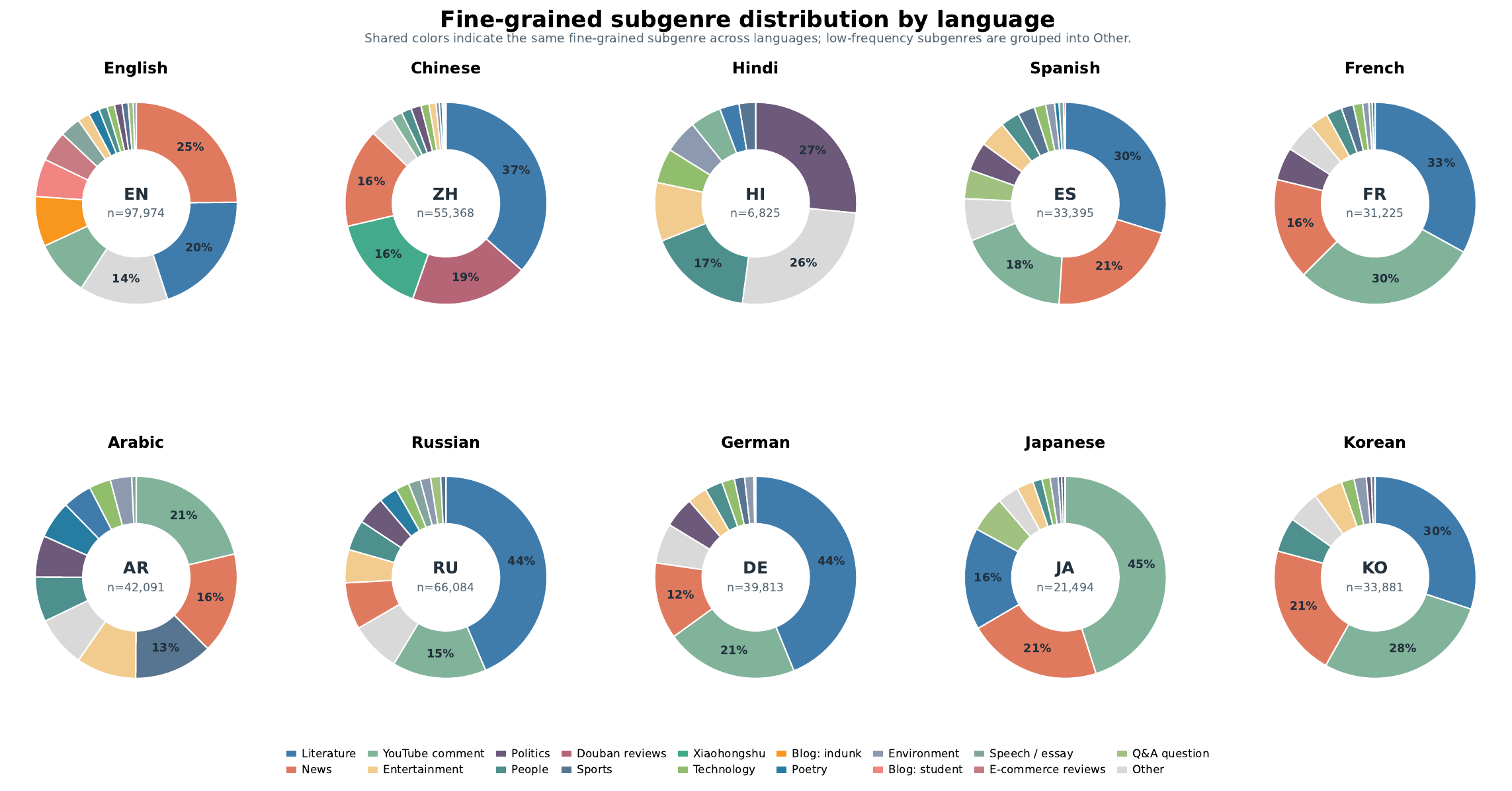}
  \caption{Fine-grained subgenre distribution by language. Each donut chart summarizes one language slice using a shared subgenre palette, while lower-frequency subgenres are grouped into \texttt{Other} to keep the figure legible.}
  \label{fig:aurabench-subgenre-by-lang}
\end{figure}

\rededit{This appendix complements the full result tables with additional dataset-profile visualizations. Figures~\ref{fig:overall-success5-bar}--\ref{fig:overall-eer-bar} summarize overall metric-level performance across models, Figure~\ref{fig:aurabench-len-by-lang} shows token-length distributions by language, and Figure~\ref{fig:aurabench-subgenre-by-lang} shows the fine-grained subgenre mix within each language.}

\section{Raw Data Sources}
\label{sec:appendix-raw-sources}

{\color{editred}AuthBench draws on $17$ publicly available input sources spanning multiple platforms, domains, and languages. Each raw item is mapped into a unified schema (Section~\ref{sec:construction}) with provenance (\texttt{source\_id}), language (\texttt{lang}), a normalized genre label (\texttt{genre}), and token length $\ell(d)\in\mathbb{N}$ computed under a fixed tokenizer. Table~\ref{tab:raw-sources} lists these sources, and Table~\ref{tab:retained-sources} reports their realized contribution to the current benchmark release.}

\paragraph{Citation policy.}
When a dataset has an associated peer-reviewed publication, we treat it as the canonical reference (e.g., MARC for Amazon Reviews Multi; Babel Briefings for multilingual headlines).
For mirrors or redistributed versions (e.g., Kaggle or Hugging Face hosting), we additionally cite the hosting page to ensure reproducibility via stable URLs and access dates.

\begin{table}[t]
\centering
\scriptsize
\setlength{\tabcolsep}{4pt}
\renewcommand{\arraystretch}{1.10}
\begin{tabularx}{\linewidth}{@{}L{2.9cm}L{0.8cm}L{2.2cm}L{2.5cm}L{2.3cm}Y@{}}
\toprule
\textbf{Source} & \textbf{Lang.} & \textbf{Primary genre(s)} & \textbf{Scale} & \textbf{Author labels} & \textbf{Reference} \\
\midrule
Exorde & Multi & Social / news / forums & 65M+/week & Yes (author hash) &
\citep{exorde_social_media_dec2024_week1} \\
Babel Briefings & 30+ & News headlines & 4.7M & Partial (publisher / org) &
\citep{leeb2024babelbriefings,babel_briefings_dataset} \\
Amazon Reviews Multi (MARC) & 6 & E-commerce reviews & 200k+/lang & Yes (reviewer IDs) &
\citep{keung2020marc,amazon_reviews_multi_kaggle} \\
Blog Authorship Corpus & en & Blogs & 681k posts & Yes &
\citep{schler2006blogauthorship,blog_authorship_corpus_hf} \\
\rev{arXiv Abstracts (metadata snapshot)} & en & Research papers & 1.7M+ & \rev{Yes (first author from metadata)} &
\citep{arxiv_bulk_data_access,cornell_arxiv_kaggle} \\
Xiaohongshu / Weibo & zh & Social media & 11k+ & Yes (user IDs) &
\citep{xiaohongshu_aigc_comments_kaggle} \\
Douban Reviews & zh & Media/book/music reviews & 13.5M & Yes (reviewer IDs) &
\citep{douban_reviews_kaggle} \\
Hindi Discourse Stories & hi & Literature (short stories) & 53 stories & Yes &
\citep{dhanwal2020hindi_discourse_modes,hindi_discourse_github} \\
Spanish PD Books & es & Literature & 300k+ texts & Yes (metadata) &
\citep{pleias_spanish_pd_books} \\
French PD Books & fr & Literature & 289k+ books & Yes (metadata) &
\citep{pleias_french_pd_books} \\
Arabic Classical Poetry & ar & Poetry & 70k poems & Yes (poet) &
\citep{arabic_poetry_kaggle} \\
Russian PD Corpus & ru & Literature/periodicals & 8.5k titles & Yes (metadata) &
\citep{pleias_russian_pd} \\
German PD Corpus & de & Literature/newspapers & 260k+ texts & Yes (metadata) &
\citep{pleias_german_pd} \\
{\color{editred}Stack Exchange Data Dump / API crawl} & {\color{editred}Multi} & {\color{editred}Q\&A} & {\color{editred}Crawl-dependent; current processed corpus 6.5k docs} & {\color{editred}Yes (user IDs)} &
\citep{stackexchange_data_dump_timeline_2025} \\
{\color{editred}Project Gutenberg} & {\color{editred}Multi} & {\color{editred}Literature / essays / speeches} & {\color{editred}Crawl-dependent; current processed corpus 23.6k docs} & {\color{editred}Yes (metadata author)} &
\citep{project_gutenberg_offline_catalogs} \\
{\color{editred}Wikisource} & {\color{editred}Multi} & {\color{editred}Literature / drama / speeches} & {\color{editred}Crawl-dependent; current processed corpus 100.8k docs} & {\color{editred}Yes (author metadata / heuristics)} &
\citep{wikimedia_wikisource_dumps} \\
{\color{editred}YouTube comments} & {\color{editred}Multi} & {\color{editred}Social media comments} & {\color{editred}Crawl-dependent; current processed corpus 104.7k comments} & {\color{editred}Yes (comment author)} &
\citep{youtube_data_api_v3_docs} \\
\bottomrule
\end{tabularx}
\caption{Raw sources used for the current AuthBench release. ``Scale'' reflects either the approximate size of the upstream source or, for crawl-dependent sources, the size of the processed corpus produced by the current collection setting before final benchmark selection.}
\label{tab:raw-sources}
\end{table}

{\color{editred}
\begin{table}[t]
\centering
\small
\setlength{\tabcolsep}{5pt}
\renewcommand{\arraystretch}{1.12}
\begin{tabular}{@{}lrr@{}}
\toprule
\textbf{Source} & \textbf{\#Docs} & \textbf{Share} \\
\midrule
Exorde & \rededit{94{,}231} & \rededit{22.0\%} \\
Wikisource & \rededit{78{,}984} & \rededit{18.4\%} \\
Babel Briefings & \rededit{73{,}676} & \rededit{17.2\%} \\
YouTube comments & \rededit{71{,}808} & \rededit{16.8\%} \\
Blog Authorship Corpus & \rededit{22{,}494} & \rededit{5.3\%} \\
Project Gutenberg & \rededit{18{,}739} & \rededit{4.4\%} \\
Russian PD Corpus & \rededit{12{,}728} & \rededit{3.0\%} \\
Douban Reviews & \rededit{10{,}424} & \rededit{2.4\%} \\
Xiaohongshu / Weibo & \rededit{8{,}869} & \rededit{2.1\%} \\
French PD Books & \rededit{8{,}761} & \rededit{2.0\%} \\
German PD Corpus & \rededit{8{,}400} & \rededit{2.0\%} \\
Spanish PD Books & \rededit{4{,}961} & \rededit{1.2\%} \\
Amazon Reviews Multi (MARC) & \rededit{4{,}924} & \rededit{1.2\%} \\
Stack Exchange Data Dump / API crawl & \rededit{4{,}651} & \rededit{1.1\%} \\
Arabic Classical Poetry & \rededit{2{,}503} & \rededit{0.6\%} \\
arXiv Abstracts (metadata snapshot) & \rededit{1{,}784} & \rededit{0.4\%} \\
Hindi Discourse Stories & \rededit{213} & \rededit{0.0\%} \\
\bottomrule
\end{tabular}
\caption{Source composition of the current combined AuthBench export. \rededit{All 17 configured sources are represented in the materialized benchmark; the table reports their realized document counts after filtering, redundancy reduction, cross-phase merge cleanup, and final selection.}}
\label{tab:retained-sources}
\end{table}
}

{\color{editred}\rededit{No configured source is entirely absent from the current combined filtered output; however, the realized source distribution is highly skewed, with Exorde, Wikisource, Babel Briefings, and YouTube comments together contributing 74.4\% of all documents.}}

\begin{table}[t]
\centering
\scriptsize
\setlength{\tabcolsep}{4pt}
\renewcommand{\arraystretch}{1.10}
\caption{Licensing / terms and release mode by source. ``Release mode'' indicates whether we redistribute normalized text (Tier A) or provide manifest-only reconstruction (Tier B). This table covers all 17 sources in the current release.}
\label{app:license-table}

\begin{tabularx}{\linewidth}{@{}L{2.9cm}L{5.2cm}L{1.3cm}Y@{}}
\toprule
\textbf{Source} & \textbf{License / terms (summary)} & \textbf{Release mode} & \textbf{Notes / pointer} \\
\midrule
Exorde & MIT (dataset card) & Tier A & Provider page explicitly states MIT. \\
Babel Briefings & CC BY-NC-SA 4.0 (dataset card) & Tier A & Attribution, non-commercial use, and share-alike obligations apply. \\
Amazon Reviews Multi (MARC) & Provider-restricted / non-permissive redistribution; dataset card does not expose a standard open license & Tier B & Safer to require reconstruction from the original release rather than redistribute normalized text. \\
Blog Authorship Corpus & No clear standardized redistribution license found in the public host materials & Tier B & Conservative manifest-only treatment is safer than direct text redistribution. \\
arXiv Abstracts (metadata snapshot) & Metadata access is available via arXiv services, but article licenses vary by paper & Tier B & We treat this as metadata-derived text and do not redistribute PDFs or assume one uniform article license. \\
Xiaohongshu / Weibo & Platform / provider terms for redistribution are unclear & Tier B & Manifest-only unless explicit redistribution permission is documented. \\
Douban Reviews & Platform / provider terms for redistribution are unclear & Tier B & Same conservative handling as other platform-hosted review data. \\
Hindi Discourse Stories & Redistribution terms are unclear from public materials & Tier B & Treat as a research-use source unless explicit permission is documented. \\
Spanish PD Books & Public-domain source texts & Tier A & Normalized text can be redistributed as public-domain material. \\
French PD Books & Public-domain source texts & Tier A & Normalized text can be redistributed as public-domain material. \\
Arabic Classical Poetry & Public host available, but redistribution terms should be verified per host copy & Tier B & Conservative default is manifest-only unless an explicit redistributable license is documented. \\
Russian PD Corpus & Public-domain source texts & Tier A & Normalized text can be redistributed as public-domain material. \\
German PD Corpus & Public-domain source texts & Tier A & Normalized text can be redistributed as public-domain material. \\
Stack Exchange Data Dump / API crawl & CC BY-SA for public user contributions (version depends on post date; newer content is CC BY-SA 4.0) & Tier A & Attribution and share-alike requirements apply to redistributed text. \\
Project Gutenberg & Mostly U.S.-public-domain texts, but Project Gutenberg license / trademark terms still apply & Tier B & Conservative manifest-only handling avoids over-claiming unrestricted redistribution across jurisdictions and formats. \\
Wikisource & Usually CC BY-SA with possible page-level public-domain variation depending on the work & Tier B & Attribution / share-alike obligations and page-level rights should be tracked before direct redistribution. \\
YouTube comments & Governed by YouTube API Terms and Developer Policies rather than a simple open-content license & Tier B & Use manifests / derived metadata only; avoid direct text redistribution by default. \\
\bottomrule
\end{tabularx}

\vspace{2mm}
\end{table}

\subsection{Data Consent}
\label{sec:checklist-d3}
AuthBench is derived from publicly accessible datasets and public-web sources released or exposed under documented provider licenses or site terms (Appendix~\ref{sec:appendix-raw-sources}).
We did not collect new data directly from individuals, and we did not contact data subjects.
Where consent mechanisms are relevant, we rely on the original dataset providers’ terms or the governing platform/site terms for public availability and redistribution.
We further reduce privacy risk by anonymizing author identifiers (Section~\ref{sec:overview}; Section~\ref{sec:construction}) and applying conservative safety filtering (Appendix~\ref{subsec:construct-filter}).



\end{document}